\pdfoutput=1

\documentclass[11pt]{article}
\usepackage[table]{xcolor}
\usepackage[]{ACL2023}

\usepackage{times}
\usepackage{latexsym}
\usepackage[T2A,T1]{fontenc}
\usepackage{highlight}
\usepackage{highlighta}
\usepackage[utf8]{inputenc}
\usepackage{dirtytalk}
\usepackage{longtable}
\usepackage{highlightuk}
\usepackage{listings}
\usepackage{multirow}
\usepackage{multirow}
\usepackage{subcaption}
\usepackage{arydshln} 
\usepackage{siunitx}
\usepackage{makecell}
\usepackage{adjustbox}
\usepackage{microtype}
\usepackage{booktabs}
\usepackage{amsmath}
\usepackage{amssymb}
\usepackage{hyperref}
\usepackage{scalerel}
\usepackage{enumitem}
\usepackage{todonotes}
\usepackage{inconsolata}
\usepackage{algorithm}
\usepackage{algpseudocode}
\usepackage{graphicx}
\usepackage{booktabs, tabularx}
\usepackage{tcolorbox}
\renewcommand{\texttt}[1]{%
  \begingroup
  \ttfamily
  \begingroup\lccode`~=`/\lowercase{\endgroup\def~}{/\discretionary{}{}{}}%
  \begingroup\lccode`~=`[\lowercase{\endgroup\def~}{[\discretionary{}{}{}}%
  \begingroup\lccode`~=`.\lowercase{\endgroup\def~}{.\discretionary{}{}{}}%
  \catcode`/=\active\catcode`[=\active\catcode`.=\active
  \scantokens{#1\noexpand}%
  \endgroup
}
\newcommand{\model}[1]{{\ttfamily #1}}

\newcommand{\cyrillic}[1]{%
        \fontencoding{T2A}\selectfont
        #1%
        \fontencoding{T1}\selectfont
}

\newcommand{\CC}[1]{%
  \raisebox{-0.3ex}{%
    \scalerel*{\includegraphics{figures//flags/#1.png}}{\scalebox{1.35}{\textbf{B}}}%
  }%
}
\newcommand{\benchmarkname}[1]{{\textsc{#1}\textsc{BORSch}}}

\title{What are Foundation Models Cooking in the Post-Soviet World? \\
}

\author{Anton Lavrouk, Tarek Naous, Alan Ritter, Wei Xu \\
  Georgia Institute of Technology \\
  \small{
 \texttt{\{antonlavrouk, tareknaous\}@gatech.edu; \{alan.ritter, wei.xu\}@cc.gatech.edu}}}

\begin{document}
\maketitle

\begin{abstract}

The culture of the Post-Soviet states is complex, shaped by a turbulent history that continues to influence current events. 
In this study, we investigate the \textit{Post-Soviet cultural food knowledge} of foundation models by constructing \benchmarkname{}, a multimodal dataset encompassing 1147 and 823 dishes in the Russian and Ukrainian languages, centered around the Post-Soviet region.
We demonstrate that leading models struggle to correctly identify the origins of dishes from Post-Soviet nations in both text-only and multimodal Question Answering (QA), instead over-predicting countries linked to the language the question is asked in.
Through analysis of pretraining data, we show that these results can be explained by misleading dish-origin co-occurrences, along with linguistic phenomena such as Russian-Ukrainian code mixing. 
Finally, to move beyond QA-based assessments, we test models' abilities to produce accurate visual descriptions of dishes. The weak correlation between this task and QA suggests that QA alone may be insufficient as an evaluation of cultural understanding. To foster further research, we will make \benchmarkname{} publicly available at \url{github.com/alavrouk/BORSch}. 

\end{abstract}

\section{Introduction}

The Post-Soviet states have long held their own cultural and linguistic identities. During the Soviet era, these identities were pressured through forced assimilation under the Russian language and culture \cite{silver1974social}. Now, 33 years after the collapse of the Soviet Union, the Post-Soviet world continues to repair the damage inflicted by this so-called \say{Sovietization} \cite{Rutland_2023}. 

\begin{figure}[t]
    \centering
    \includegraphics[width=\linewidth]{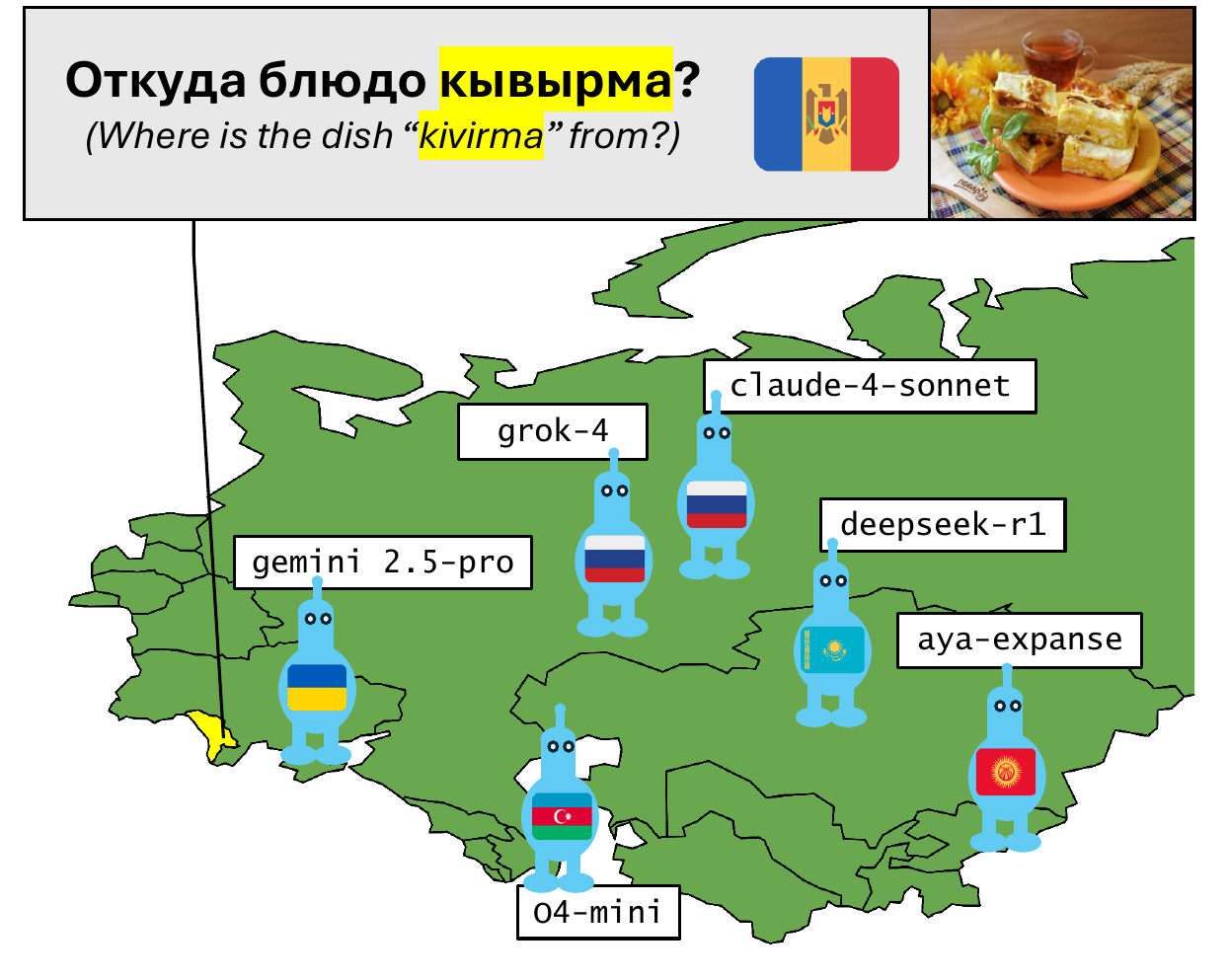}
    \caption{A map of predictions from various popular consumer-facing models for the dish \cyrillic{кывырма} (\textit{kivirma}), a traditional pastry from the Moldovan region of Gagauzia, which is home to a significant Russian-speaking population. The models were prompted in Russian, with the English translation also shown.}
    \vspace{-0.2in}
    \label{fig:example}
\end{figure}

As foundation models continue to gain prominence, it is important that they are able to represent each Post-Soviet state. 
Yet, when examined via \textit{food dishes}, an important element in every culture \cite{anderson2014everyone}, we find that they lack crucial knowledge. 
For example, \cyrillic{кывырма} (pronounced \say{ky-vyr-MA}) is a dish from Gagauzia, a region of Moldova where Russian is commonly spoken \cite{mayer2014gagauz}. 
However, Figure~\ref{fig:example} shows that when asked in Russian, multilingual models fail to identify the origins of this dish, with each one predicting an incorrect Post-Soviet nation.

In order to further investigate these deficiencies, we conduct a detailed exploration of food culture understanding in foundation models across both text and image modalities. We focus our study on the Russian and Ukrainian languages, analyzing how the push of Ukrainian \say{de-Sovietization} \cite{boman2023coexistence} and the pull of historical Russian interference on Ukrainian culture \cite{boychuk2023effect} impacts the cultural perceptions of foundation models. Overall, our contributions are as follows:

\begin{itemize}[noitemsep, topsep=0pt, parsep=0pt, partopsep=0pt]

    \item We construct \benchmarkname{}\footnote{Named after the famous Ukrainian dish, \cyrillic{борщ}.} (\textbf{B}enchmark \textbf{O}f \textbf{R}egional di\textbf{S}hes), a dataset for evaluating foundation models on multimodal food culture understanding in Russian and Ukrainian (\S\ref{sec:dataset}). \benchmarkname{} is constructed via a bootstrapped entity extraction approach for collecting culturally relevant food dishes from web crawl corpora, with human-in-the-loop validation (\S\ref{subsec:bootstrapping}).

    \item To compare models prompted in Russian and Ukrainian, we perform text and vision country of origin Question Answering (QA) using dishes in \benchmarkname{}. Beyond varying performances across countries, we find that models prompted in these languages over-predict their respective countries of origin (\S\ref{subsec:parallel_qa}). 

    \item To gain a more nuanced understanding of how models perform on Post-Soviet dishes in Ukrainian, we examine how the Russian-Ukrainian pidgin\footnote{A pidgin is a simplified language for communication between people with different native tongues \cite{romaine2017pidgin}.} \textit{surzhyk} influences Ukrainian QA and VQA performance (\S\ref{subsec:surzhyk}).
    
    \item Through pretraining data analysis, we find many instances where \benchmarkname{} dishes co-occur with non-origin countries, harming model QA performance. In contrast to English corpora, these issues in Russian and Ukrainian largely stem from poor web-scraping (\S\ref{subsec:pretraining}).

    \item Finally, we conduct an experiment which queries models for descriptions of dishes in \benchmarkname{}, which we then evaluate using a modality transition from text to image. We find this to be a challenging task with limited correlation to QA experiments (\S\ref{sec:descriptions}). 
\end{itemize}

\section{Related Work}
\label{sec:related_work}

\paragraph{Cultural Knowledge Bases.} 
Recent interest regarding cultural-knowledge in foundation models has led to numerous studies attempting to quantify it \cite{hershcovich2022challengesstrategiescrossculturalnlp, adilazuarda-etal-2024-towards, liu2024culturallyawareadaptednlp}. Some studies construct multilingual knowledge bases of cultural assertions (e.g., \textit{\say{In Bhutan, there is a tradition of wearing "Khyenkhor Robes" woven with threads infused with blessings from Buddhist monks}}) \cite{Nguyen_2023, Nguyen_2024, fung2024massivelymulticulturalknowledgeacquisition}. Other works craft benchmarks of culturally-specific questions (e.g., \textit{\say{What is the story of the series Al-Manassa?}}) \cite{yin2022geomlama, myung2024blendbenchmarkllmseveryday, shen-etal-2024-understanding, arora2024calmqaexploringculturallyspecific}. Further research expands on such directions to support multimodality \cite{ramaswamy2023geodegeographicallydiverseevaluation, liu2023cultural, libovický2025cusqalocalknowledgeorientedopenendedquestion}. There are also additional studies which focus exclusively on vision-based tasks such as culturally informed image generation \cite{bhatia2024local, karamolegkou2024vision, kannen2024beyond}, visually grounded reasoning \cite{schneider2024m}, and image transcreation \cite{khanuja2024image}. While some of these works include food as part of their overall assessment, they mainly focus on broad cultural understanding. Meanwhile, we offer a more in-depth analysis on the nuances of cultural \textit{food} knowledge.

\paragraph{Cultural Food Knowledge.}
Food knowledge is a key element of culture, and is thus frequently evaluated in foundation models. Some studies assess model comprehension of culinary practices or dishes through pragmatic questioning (e.g., "\textit{While eating, when does one drink Cantonese seafood soup?}") \cite{palta2023fork, yao2023benchmarking, putri2024can, li2024foodieqamultimodaldatasetfinegrained}. Another line of reasoning uses food to attribute cultural generations to pretraining data \cite{li2024attributingcultureconditionedgenerationspretraining}.
Finally, a group of works measures food culture understanding in foundation models by testing them on a culturally diverse set of food-dish entities. However, the food-dishes used for evaluating models in past work are obtained solely from either Wikidata \cite{zhou2024does} or Wikipedia \cite{winata2024worldcuisines}, which we show leads to missing out on many culture-specific dishes in non-English languages (\S\ref{subsec:wikidata}). For example, the food-dishes originating from Russia and Ukraine in \textsc{WorldCuisines} \cite{winata2024worldcuisines} cover only 20.8\% of the dishes originating from Russia and Ukraine that we provide in \benchmarkname{}.


\begin{figure*}[t]
    \centering
    \begin{subfigure}[t]{0.333\linewidth}
        \centering
        \includegraphics[width=\linewidth]{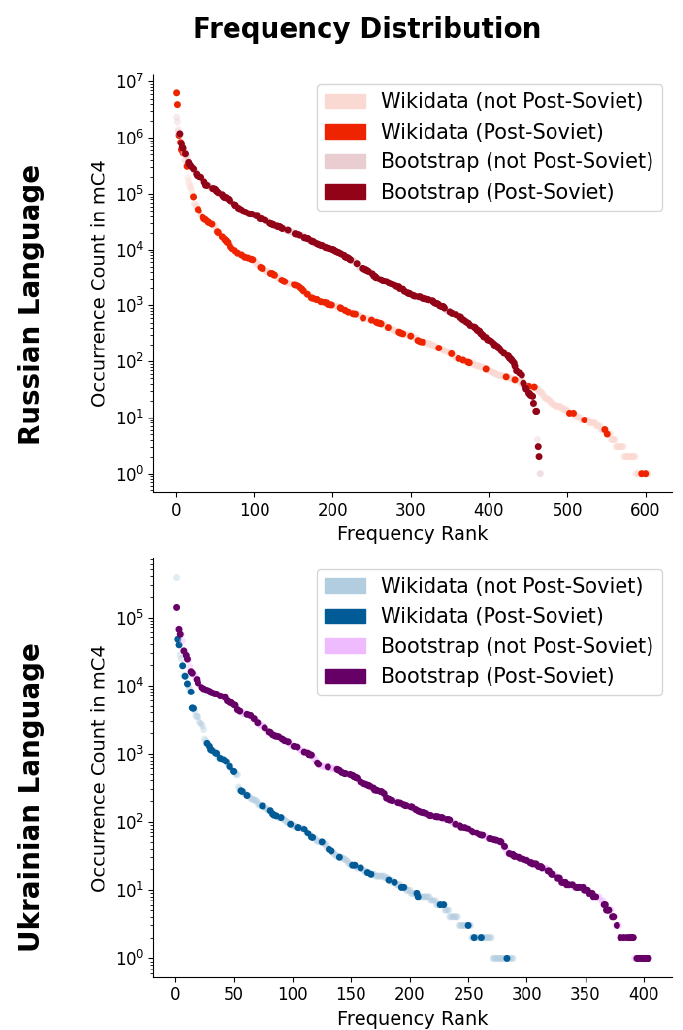}
        \caption{}
        \label{fig:figure2a}
    \end{subfigure}%
    \hfill
    \begin{subfigure}[t]{0.666\linewidth}
        \centering
        \includegraphics[width=\linewidth]{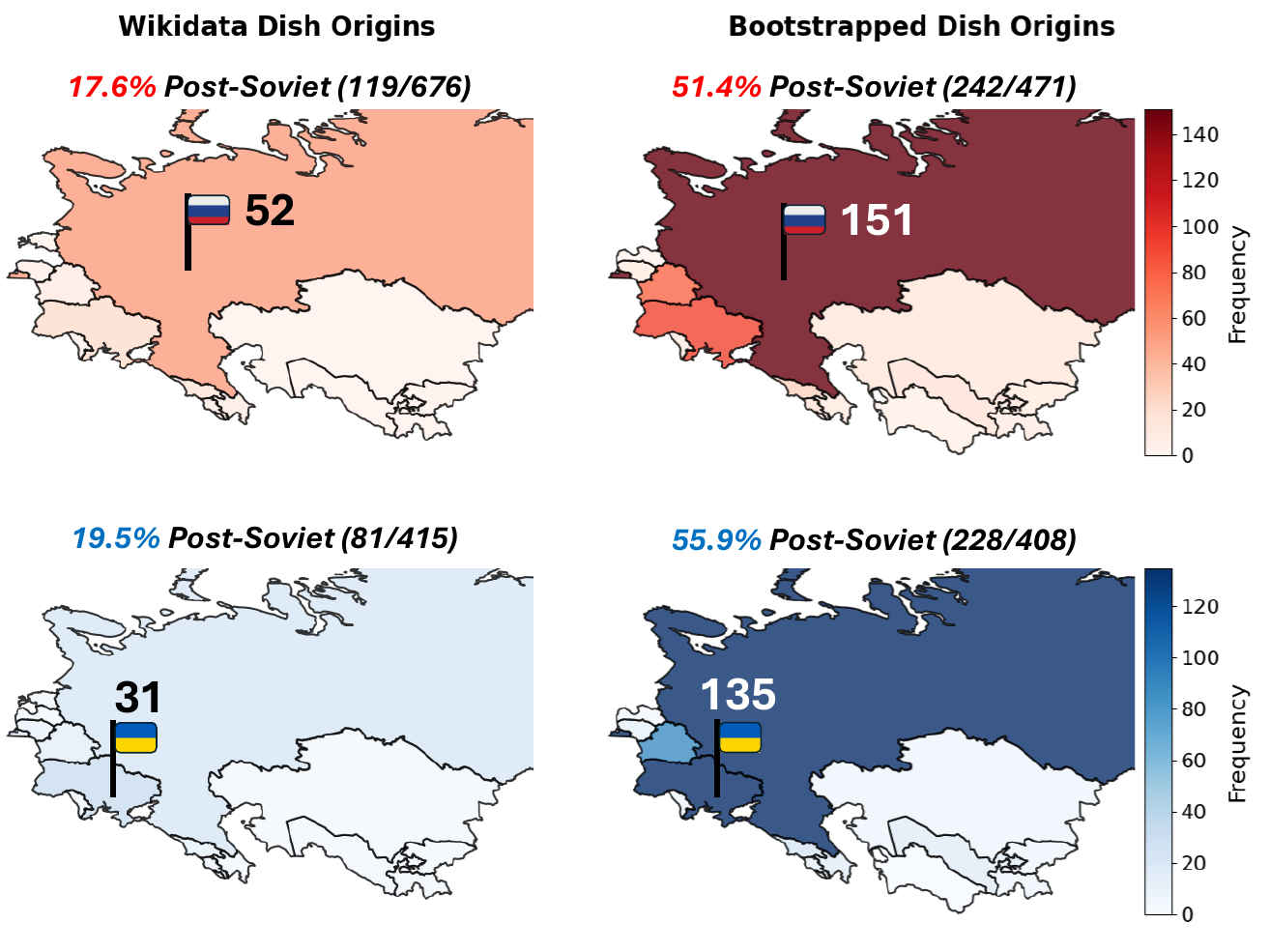}
        \caption{}
        \label{fig:figure2b}
    \end{subfigure}
\caption{
        \textbf{(a)} Frequency in \model{mC4} vs frequency rank in \benchmarkname{}. Culturally relevant bootstrapped dishes are both common and long-tail, while Wikidata dishes are less frequent overall. \textbf{(b)} Countries of origin of dishes in \benchmarkname{}, which were obtained from multilingual Wikidata  (\S\ref{subsec:wikidata}) and commonly used web-crawled corpora (\S\ref{subsec:bootstrapping}). While there are less bootstrapped dishes, they are more likely to originate from a Post-Soviet nation.
    }
    \vspace{-0.3cm}
    \label{fig:map_comparisons}
\end{figure*}

\paragraph{Russian and Ukrainian Culture in LLMs.} 
Existing cultural studies on the Russian language in foundation models focus on social/gender biases \cite{grigoreva2024rubia, km, li2024uncoveringdifferencespersuasivelanguage} or image generation \cite{vasilev-etal-2025-ruscode}. For the Ukrainian language, \citet{kharchenko2024llmsrepresentvaluescultures} explores cultural values of foundation models. From our understanding, our study is the first to perform a large scale exploration of Post-Soviet cultural knowledge using the Russian and Ukrainian languages in parallel.

\section{Constructing \benchmarkname{}}
\label{sec:dataset}

To enable a more in-depth assessment of models' understanding of Russian and Ukrainian food cultures, we focus on collecting both the popular and less commonly known dishes relevant to those cultures. We achieve this by first extracting all available dishes in Wikidata (\S\ref{subsec:wikidata}), then expanding on this initial set through a bootstrapped extraction approach from web-crawled data with human-in-the-loop (\S\ref{subsec:bootstrapping}). We also annotate countries of origin (\S\ref{subsec:annotation}), collect images for the dishes (\S\ref{subsec:images}), and create a Post-Soviet, parallel sub-dataset of dishes (\S\ref{subsec:parallel_construction}), enabling many multimodal evaluations.

\subsection{Extracting Food Entities from Wikidata}
\label{subsec:wikidata}
As a starting point, we acquire an initial list of dishes from the Wikidata multilingual knowledge base\footnote{\url{www.wikidata.org/wiki/Wikidata:Main_Page}} in the Russian and Ukrainian languages. 
We extract all entities that are registered under the class \say{\texttt{food}} in Wikidata, which encompasses many food-related sub-classes (e.g., \textit{sweets}, \textit{fast food}, etc.). We then manually select culturally relevant food dishes that are attributable to a specific country/countries of origin, and discard beverages and more generic food entities (e.g., globally common dishes such as \textit{grilled chicken}, branded goods such as \textit{kitkat}, etc.). The resulting coverage of food entities in Wikidata is relatively poor for both Russian (676 dishes) and Ukrainian (415 dishes) languages. 
Moreover, only 119 dishes (17.6\%) in Russian Wikidata and 81 dishes (19.5\%) in Ukrainian Wikidata are associated with origins in any Post-Soviet state, which indicates that the existing coverage is not only sparse but also lacks cultural relevance. We address this multilingual coverage gap in Wikidata by collecting additional dishes from web-crawl data.

\subsection{Bootstrapped Extraction from Corpora}
\label{subsec:bootstrapping}

Previous research by \citet{naous-etal-2024-beer} demonstrated that culturally-relevant food dishes can be collected from large web-crawl corpora. Their approach relies on extracting unigrams and bigrams appearing after a set of manually crafted patterns which likely occur before the mention of a food dish (e.g., \textit{recipe of \rule{0.65cm}{0.15mm}}, \textit{how to cook \rule{0.65cm}{0.15mm}} , etc.). This was followed by human annotation to filter out erroneous extractions. We build on this method by performing bootstrapped pattern-based extraction with a human-in-the-loop to iteratively collect food dishes from the Russian and Ukrainian portions of the \model{mC4} web-crawl corpus \cite{xue-etal-2021-mt5}.
  
We start with the dishes obtained from Wikidata (\S \ref{subsec:wikidata}) as a seed list, which we use to search the corpus of each respective language and retrieve all 3-gram and 4-gram patterns that precede any dish.
We then ask a human annotator to select five from the 100 most frequent patterns.
Using the selected patterns, we search the corpus again and extract every unigram that appears after a 3-gram pattern and every bigram that appears after a 4-gram pattern.
This strategy exploits the exponential growth in n-gram counts: rare 4-gram patterns limit bigram extraction volume to make manual review more feasible, while more common 3-gram patterns ensure sufficient unigram extraction volume.

Finally, we de-duplicate the extracted unigrams and bigrams, which results in up to 10k extractions that we give to human annotators to manually filter for food dishes. We repeat this bootstrapping process for two more rounds. Detailed statistics regarding extractions during the bootstrapping process are located in Appendix~\ref{appendix:bootstrap_stats}. 
During the filtering process, a random sample of 3770 extractions (2015 in Russian and 1755 in Ukrainian) underwent double annotation, yielding substantial annotator agreement with Cohen's Kappa ($\kappa$) values of 0.73 and 0.77 for Russian and Ukrainian respectively.

\subsection{Determining a Dish's Country of Origin}
\label{subsec:annotation}

In order to enable the evaluation of models' food culture understanding, we annotate each collected dish for its associated country/countries of origin. Two college educated annotators, one fluent in Russian and one fluent in Ukrainian, conducted independent research using web resources on each dish and manually labeled each dish's country of origin. In cases where dishes were found to have multiple countries of origin (20\% of dishes in Russian, 27\% in Ukrainian), particularly for areas that predate modern country borders, annotators were asked to label all relevant countries. An example is \cyrillic{чак-чак} (\textit{chak-chak}), a popular cake originating from Central Asia which existed before the Soviet Union. Its origins were labeled as Russian (Tatar and Bashkir)\footnote{These are two minority ethnic groups in Russia.}, Kazakh, Tajik, Kyrgyz, and Uzbek as it is a common delicacy in all of those nations. 

Figure~\ref{fig:figure2b} compares the origins of food dishes extracted from Wikidata (\S \ref{subsec:wikidata}) vs. the bootstrapped process (\S \ref{subsec:bootstrapping}) on the map. We find that bootstrapping retrieves more dishes that are common in the Post-Soviet region. Furthermore, as shown in Figure~\ref{fig:figure2a}, the bootstrapping process helps cover Post-Soviet dishes in Russian and Ukrainian that are both highly frequent and long-tail in corpora, while Post-Soviet dishes obtained from Wikidata consist of mostly popular dishes. 

\subsection{Dish Image Collection}
\label{subsec:images}

To facilitate vision-languages analyses, we collect up to 5 images for each dish in \benchmarkname{}. We first searched for images in Wikimedia Commons\footnote{\url{commons.wikimedia.org/wiki/Main_Page}}, a collection of freely usable media files. This process enabled image retrieval for 74\% of dishes in Russian and 68\% in Ukrainian. For the remainder of the dishes which did not have images in Wikimedia Commons, we used the Google Custom Search API\footnote{\url{developers.google.com/custom-search/v1/}} and queried for images with a Creative Commons (CC) open license. All retrieved images were then manually filtered to remove irrelevant content (see filtering interface in Appendix~\ref{appendix:image_annot}). In total, we collected 5285 images in Russian (3.64 images per dish on average), and 2907 images in Ukrainian (3.53 images per dish on average). Additionally, we inspect potential VLM pretraining data contamination involving these images in Appendix~\ref{appendix:contamination}.

 \begin{figure*}[t]
    \centering
    \includegraphics[width=\linewidth]{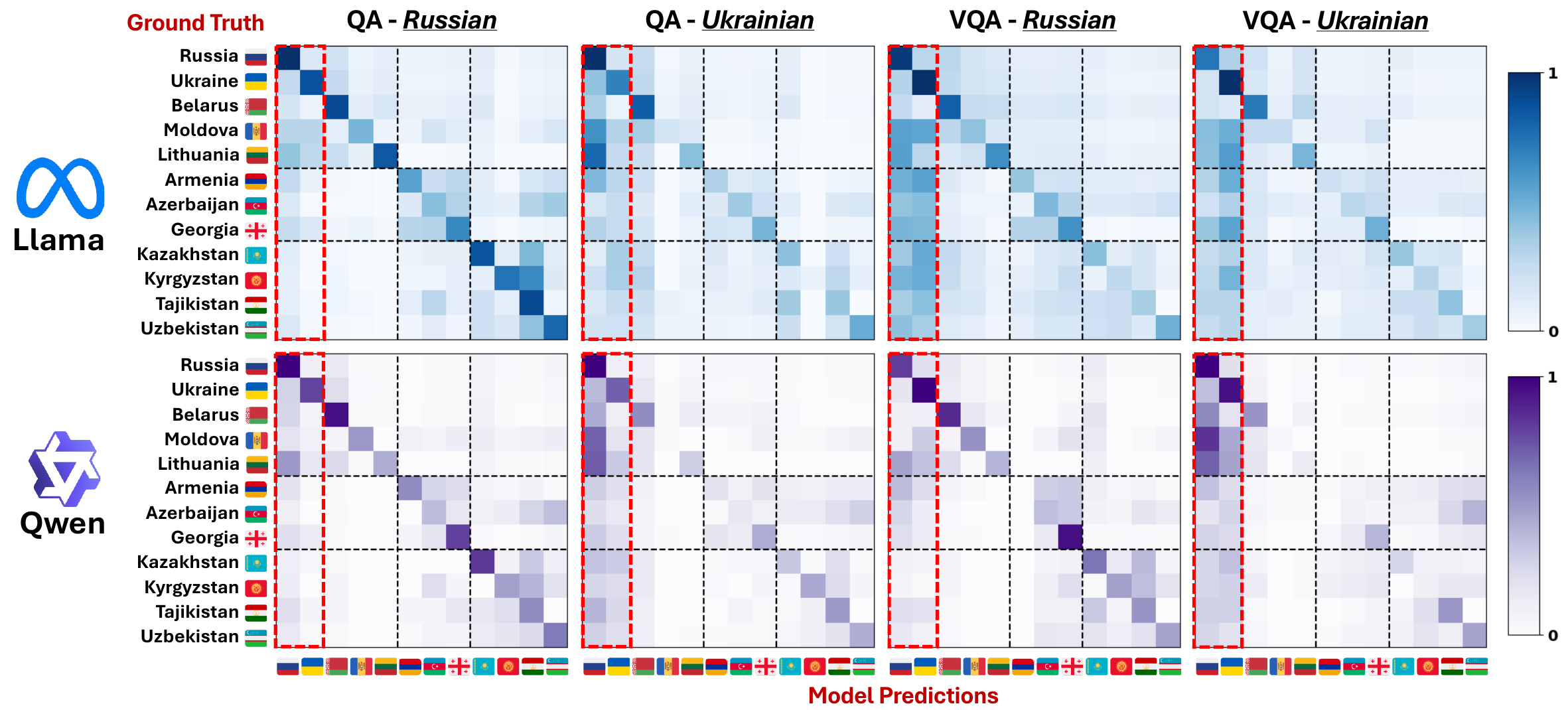}
    \caption{Confusion matrices for country-of-origin QA/VQA on dishes in \benchmarkname{}. Models exhibit a low recall on \CC{RU} Russian and \CC{UA} Ukrainian dishes, and struggle with Post-Soviet countries in the Caucasus (\CC{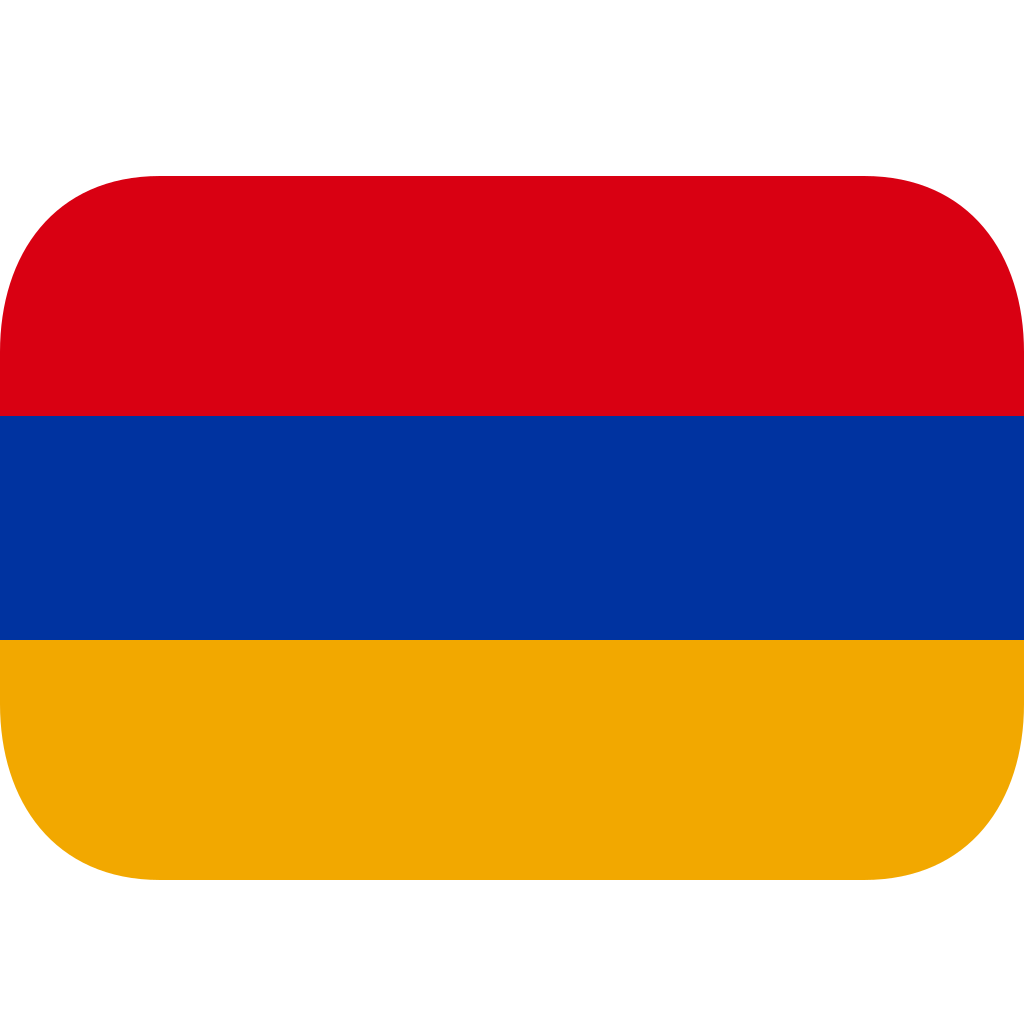} Armenia, \CC{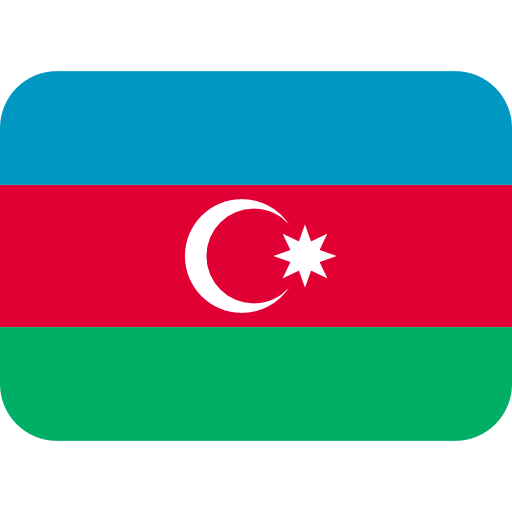} Azerbaijan, \CC{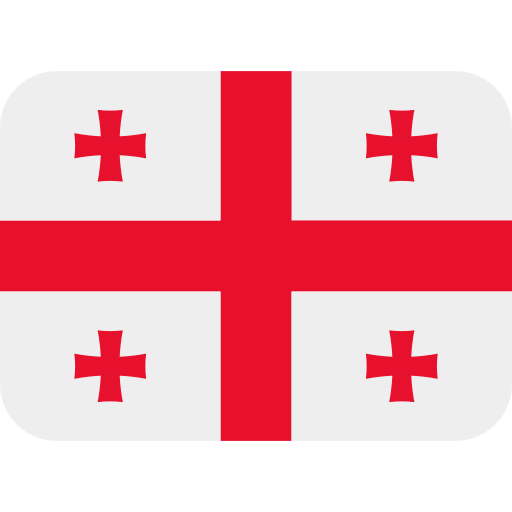} Georgia) and Central Asia (\CC{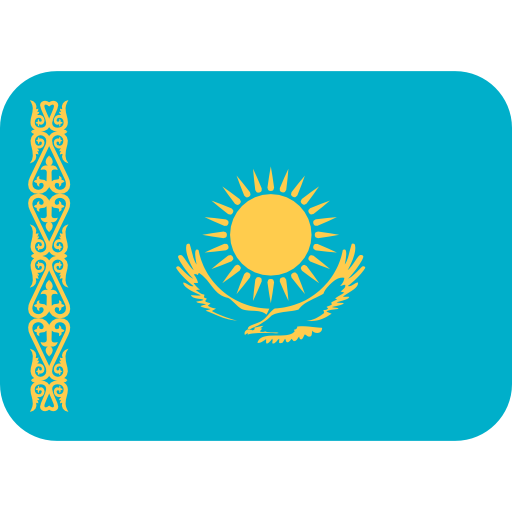} Kazakhstan, \CC{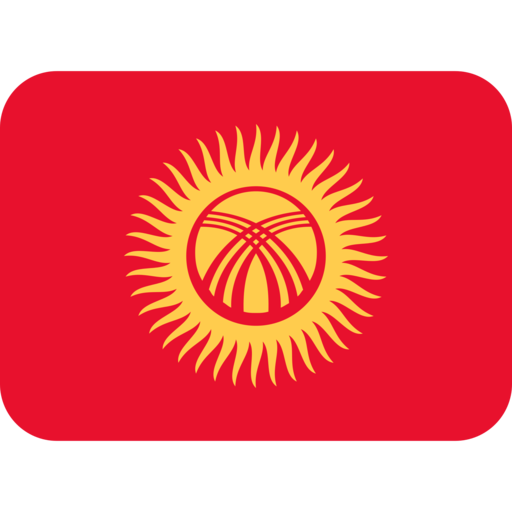} Kyrgyzstan, \CC{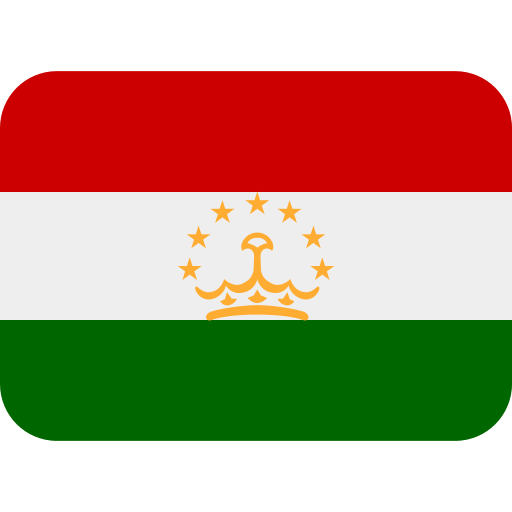} Tajikistan, \CC{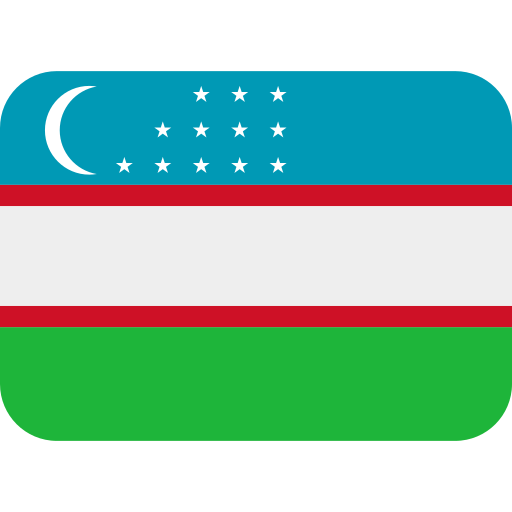} Uzbekistan). \CC{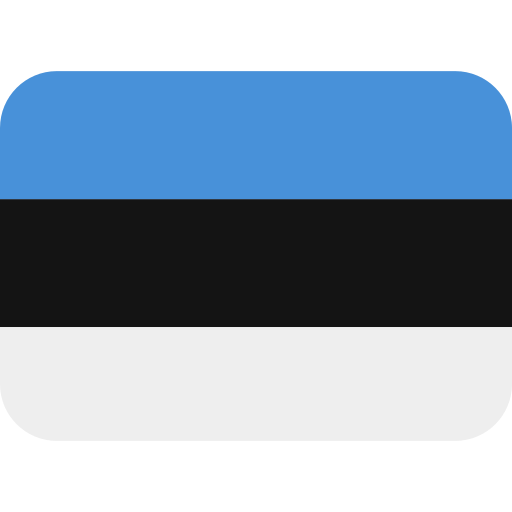} Estonia, \CC{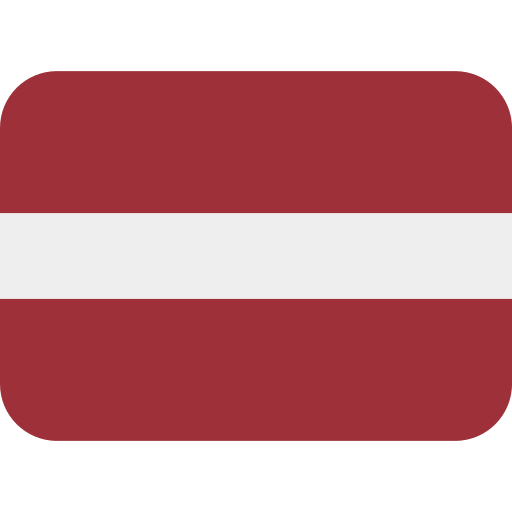} Latvia, and \CC{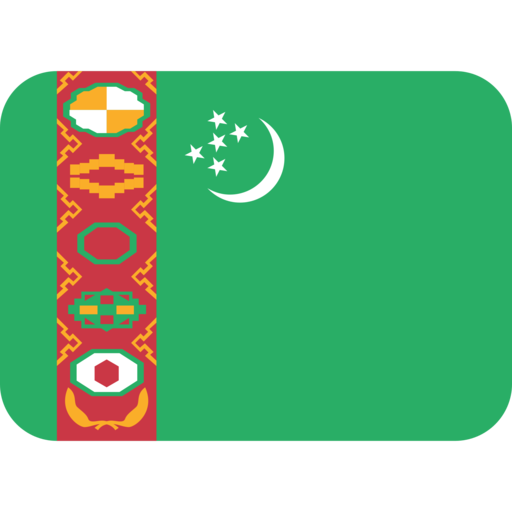} Turkmenistan are excluded due to low sample size, and the confusion matrices are row (truth) normalized, as each country is not equally represented in \benchmarkname{}.}
    \vspace{-0.4cm}
    \label{fig:qa_and_vqa}
\end{figure*}

\subsection{Aligning Russian \& Ukrainian \benchmarkname{}}
\label{subsec:parallel_construction}
To enable direct comparisons between languages, we manually translate (and transliterate, when necessary) dishes originating from Post-Soviet countries in the Russian set to Ukrainian and vice-versa, resulting in a parallel sub-dataset of 433 dishes with names in both Russian and Ukrainian. Of these dishes, 174 (40\%) appear in both datasets, while 126 (29\%) are unique to the Russian dataset and 133 (31\%) are unique to the Ukrainian dataset. The distribution of origins for the parallel sub-dataset can be found in Appendix~\ref{appendix:parallel_dist}. We validated the dish origin annotations in this parallel sub-dataset by engaging a second annotator, achieving substantial agreement with a Cohen's Kappa ($\kappa$) of 0.879. Furthermore, we note that each dish in this sub-dataset can now have up to 10 images (if it was originally a part of both Russian and Ukrainian \benchmarkname{}). 

\section{LLM Performance: Dish QA \& VQA}
\label{sec:four}

To begin, we test foundation models prompted in Russian and Ukrainian on QA and VQA tasks focusing on dish origins within the parallel sub-dataset of \benchmarkname{} (\S\ref{subsec:parallel_qa}). Then, to further understand these results, we explore the effect of Russian code mixing on Ukrainian QA and VQA (\S\ref{subsec:surzhyk}). Finally, we investigate dish-country co-occurrences in the underlying pretraining data as a factor influencing both Russian and Ukrainian QA (\S\ref{subsec:pretraining}).

\subsection{Parallel Country of Origin QA \& VQA}
\label{subsec:parallel_qa}
We first assess a model's ability to predict a dish's country/countries of origin in two setups: \textbf{(i)} standard text-based Question Answering (QA) where the model is provided the dish name and \textbf{(ii)} Visual Question Answering (VQA) where the model is provided an image of the dish. We use the \textbf{Post-Soviet parallel sub-dataset} of \benchmarkname{}, enabling fair, cross-lingual comparisons.

\paragraph{Setup.}
We evaluate \model{Qwen2-72B-Instruct} \allowbreak\cite{qwen2} and \model{Llama-3.1-70B-Instruct} \cite{dubey2024llama} on text-only QA tasks, and their vision-enabled counterparts \model{Qwen2\allowbreak-VL-72B-Instruct} and \model{Llama-3.2-90B-Vision\allowbreak-Instruct} for VQA tasks\footnote{More information in Appendix \ref{appendix:model_choice} regarding model choice.}. We prompt these models with open-ended questions to align with real-world applications \cite{rottger-etal-2024-political}. To identify countries in each model's response, we use \model{spaCy}'s multilingual NER tool, known to be highly effective for location recognition \cite{Honnibal_spaCy_Industrial-strength_Natural_2020}. After extracting named entities, we search them for country names or aliases (e.g., \textit{Czech Republic} vs. \textit{Czechia}). This two-step approach allows us to detect any missing aliases by manually analyzing named entities not in our country/alias list.
Models are prompted using five different variants of the same question, each containing placeholders for dish names (see Appendix~\ref{appendix:prompts}). We attach all (up to 10) available images to each VQA prompt. 

\begin{figure}[t]
    \centering
    \includegraphics[width=\linewidth]{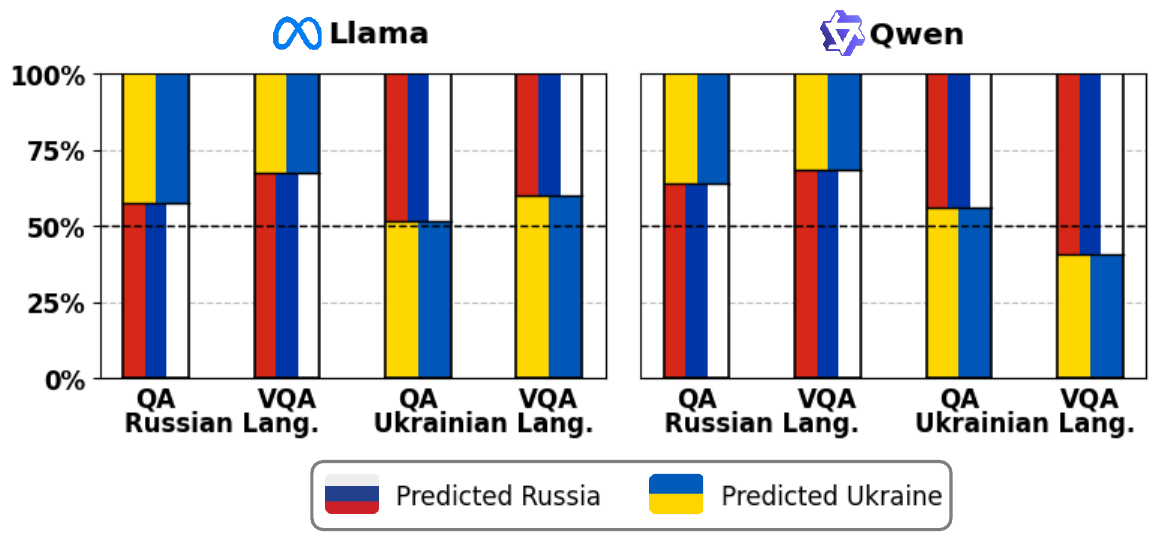}
      \caption{{For dishes originating from both \CC{RU} Russia and \CC{UA} Ukraine, Russian models tend to predict Russia more often than Ukraine, and Ukrainian models tend to predict Ukraine more often than Russia. The proportions shown exclude non-Russian/Ukrainian predictions.}}
    \vspace{-0.4cm}
    \label{fig:pies}
\end{figure}

\paragraph{Results.} Confusion matrices\footnote{As models can generate multiple predictions, we utilize the methodology in \citet{heydarian2022mlcm} and \citet{krstinic2020multi} to construct \say{multi-label confusion matrices.}} for the country-of-origin QA/VQA experiments are presented in Figure~\ref{fig:qa_and_vqa}. We focus on Post-Soviet predictions for cultural specificity, with frequent non-Post-Soviet errors listed in Appendix~\ref{appendix:incorrect_preds}.
Furthermore, in Appendix~\ref{appendix:vqa_with_name}, we report the results of a modified VQA task where models are additionally provided dish names alongside images. This additional information improves performance on most dishes.
Overall, we find that when prompted in Russian and Ukrainian, models frequently over-predict Russia and Ukraine as dish origins. In Figure~\ref{fig:qa_and_vqa}, this is evident from the wide distributions in the Russian and Ukrainian prediction columns. We investigate this further by focusing on dishes whose origins include both Russia and Ukraine (35\% of the parallel corpus). Figure~\ref{fig:pies} shows that when prompted in Russian, models are more likely to predict Russia as an origin for these dishes, while models prompted in Ukrainian are more likely to predict Ukraine. This mainly impacts models prompted in Russian and affects QA tasks more than VQA. 

\subsection{Impact of Code Mixing on QA \& VQA} 
\label{subsec:surzhyk}

One potential factor influencing model QA/VQA performance in Ukrainian is Russian code mixing. In the period following Ukraine's establishment as a sovereign state in 1991, many Russian speakers in Ukraine transitioned to speaking the Ukrainian language \cite{fomenko2023brand}.
Understandably, this linguistic adjustment was not instant, which is why \textit{surzhyk}, referring to various mixed Russian-Ukrainian codes, has since gained a foothold in Ukraine \cite{doi:10.1177/00220221241256322}. For example, in surzhyk, sentences that are otherwise fully Ukrainian often incorporate Russian words, particularly nouns \cite{podolyan2005ukrainians}. On the other hand, the reverse phenomenon is far less common. Using \model{mC4} as a representative corpus of text data (more on this later), we find that Ukrainian code mixing accounts for 17\% of \benchmarkname{} dish occurrences in Russian \model{mC4}, which is far less than Russian dish occurrences in Ukrainian \model{mC4} (41\%).

To study surzhyk in pretraining corpora, we use \benchmarkname{}, particularly since the dish names in our parallel Ukrainian-Russian sub-dataset only differ by an average edit distance of 2.3 characters. This similarity makes the dishes in \benchmarkname{} especially susceptible to code mixing (e.g., Ukrainian \cyrillic{пир}\textbf{i}\cyrillic{г} versus Russian \cyrillic{пир}\textbf{o}\cyrillic{г})\footnote{Directly translates to \say{pie,} but colloquially represents a specific class of Eastern European pastries.}.

\paragraph{Setup.}
Using the Russian and Ukrainian dish names in the parallel sub-dataset of \benchmarkname{} (\S~\ref{subsec:parallel_construction}), we search for instances of Russian code-mixing in Ukrainian corpora. In particular, we search \model{mC4}, the most widely used and extensively studied open multilingual corpus for pretraining \cite{kreutzer2022quality}. While more recent corpora contain newer, higher quality data, they do not contain multilingual components, which are crucial for our analysis.
We then quantify surzhyk by analyzing the \textit{difference in mC4 occurrences of the Ukrainian dish name and the Russian dish name}.

To assess model performance, we calculate the Jaccard score per dish as each dish may have a varied number of countries as its origin. The Jaccard score \cite{jaccard1901etude} measures the similarity between the predicted and ground truth country of origin sets as the size of their intersection divided by the size of their union. Importantly, this \textit{set overlap} Jaccard score accounts for countries predicted by the model which are not part of the gold set.
Additionally, to enable the analysis of surzhyk in VQA, we modify the VQA experiment by asking the model to \textit{name} a dish given its image. We evaluate model performance on this experiment using \textit{exact match accuracy over 5 prompts} with a tolerance of 1 edit (Levenshtein) distance. While exact match is a common QA metric \cite{rajpurkar2016squad}, we introduce the 1 edit distance tolerance due to Russian and Ukrainian noun declension \cite{press2015ukrainian, comrie2018russian}. Full, per-country results for this experiment can be found in Appendix~\ref{appendix:dish_vqa}.

\begin{figure}[t]
    \centering
    \includegraphics[width=0.975\linewidth]{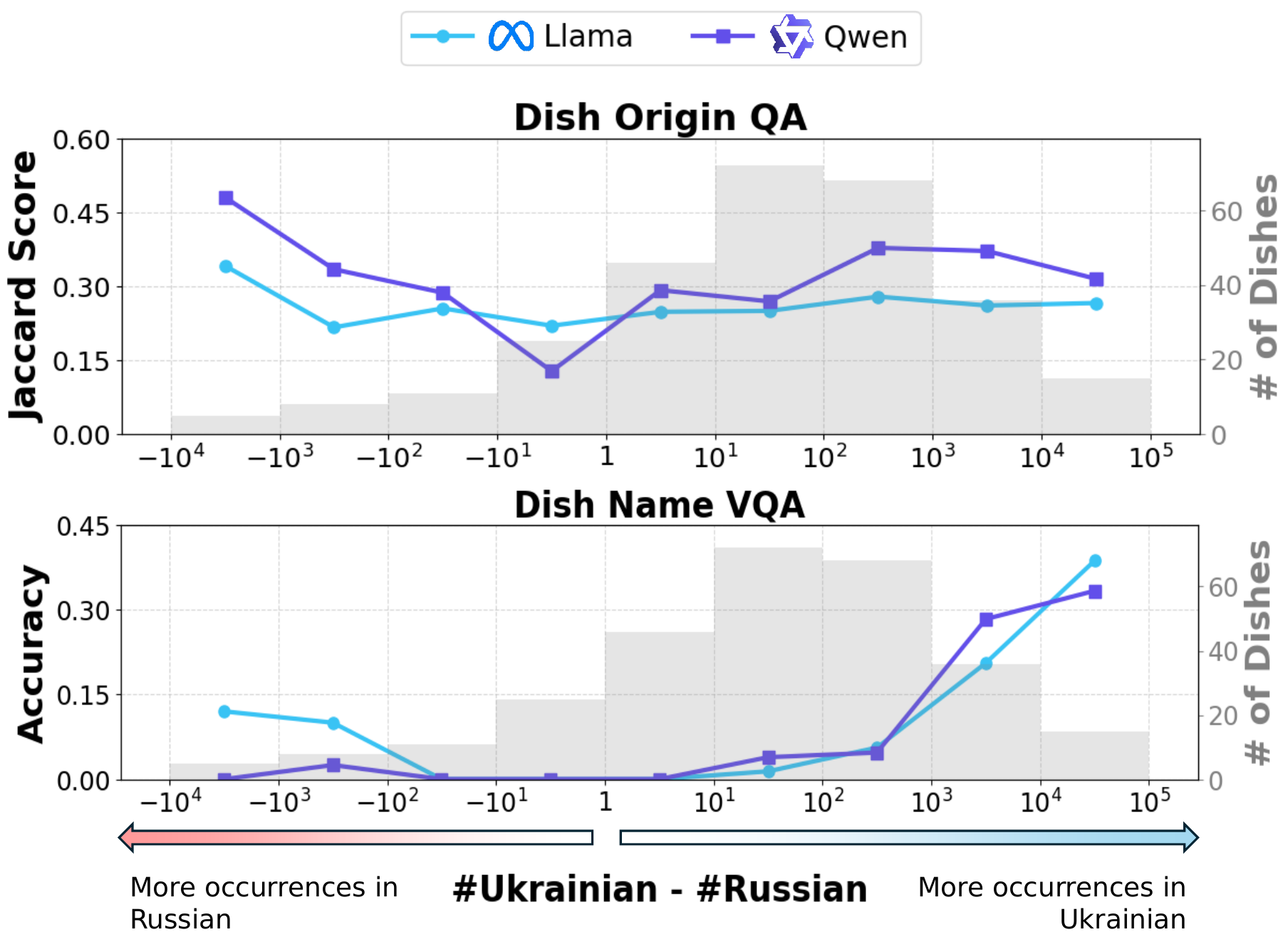}
    \caption{Dish origin QA and dish name VQA tasks suffer when prompting Qwen and Llama in Ukrainian because dish names lack standardization in Ukrainian corpora, which can occur due to the use of \textit{surzhyk} (a mixed Russian-Ukrainian code). Dishes with identical Russian and Ukrainian names are not analyzed.}
    \vspace{-0.4cm}
    \label{fig:contamination}
\end{figure}

\paragraph{Results.}  
We begin by calculating the average QA Jaccard score and VQA exact match accuracy across nine log-spaced bins. These bins are determined by occurrence differences in Ukrainian mC4 (\#Ukrainian - \#Russian) for the 66.5\% of dishes in the parallel dataset that have distinct Russian and Ukrainian names. The results are located in Figure~\ref{fig:contamination}. 
For dish origin QA, we observe the poorest performance when the pretraining data contains an approximately equal mix of Russian and Ukrainian names, while dishes standardized to either a Russian or Ukrainian name perform better. 
Similarly, for dish name VQA, we find that it is important for the dish name to be standardized, although preferably using the Ukrainian name.

\subsection{Impact of Co-Occurrences on QA}
\label{subsec:pretraining}

Previously, we demonstrated that QA performance in Ukrainian can be affected by Russian code mixing (\S\ref{subsec:surzhyk}). We now turn our attention to \textbf{incorrect dish-country co-occurrences} in both Russian and Ukrainian pretraining data.

\paragraph{Setup.} To start, we analyze how frequently each \benchmarkname{} dish (full dataset) appears alongside country names and aliases in the Russian and Ukrainian subsets of \model{mC4}. 
For each dish, we define two metrics. The \textbf{correct co-occurrence count} is the number of \model{mC4} documents that mention both the dish and its true country of origin. The \textbf{correct co-occurrence ratio} divides this count by the total mentions of that dish with any country (correct or incorrect). We then examine how these metrics impact model country of origin QA performance.

\begin{figure}[t]
    \centering
    \includegraphics[width=\linewidth]{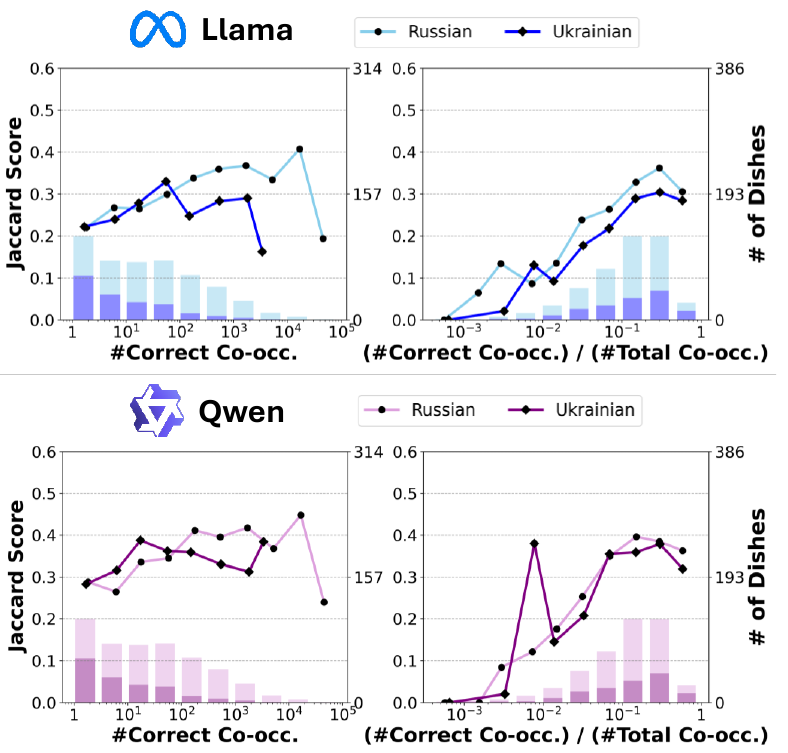}
    \caption{While a higher number of correct dish-country co-occurrences (\model{\#Correct Co-occ.}) supports better text-only QA performance, the ratio of correct v.s. total co-occurrences (\model{\#Correct Co-occ./\#Total Co-occ.}) proves even more crucial in both Llama and Qwen.}
    \vspace{-0.2cm}
    \label{fig:freq_coc}
\end{figure}

\paragraph{Results.} Figure~\ref{fig:freq_coc} shows text-only QA performance for Qwen and Llama averaged over dishes which are grouped into 10 log-spaced bins based on their correct co-occurrence counts or ratios. We find an improvement in text-only QA across both Russian and Ukrainian languages as dishes occur more frequently with the correct country in pretraining data. More importantly, the improvement in Jaccard score is steeper when looking at the correct co-occurrence ratio, indicating that it is critical for a dish not to co-occur frequently with incorrect countries of origin.

\begin{figure}[t]
    \centering
    \includegraphics[width=\linewidth]{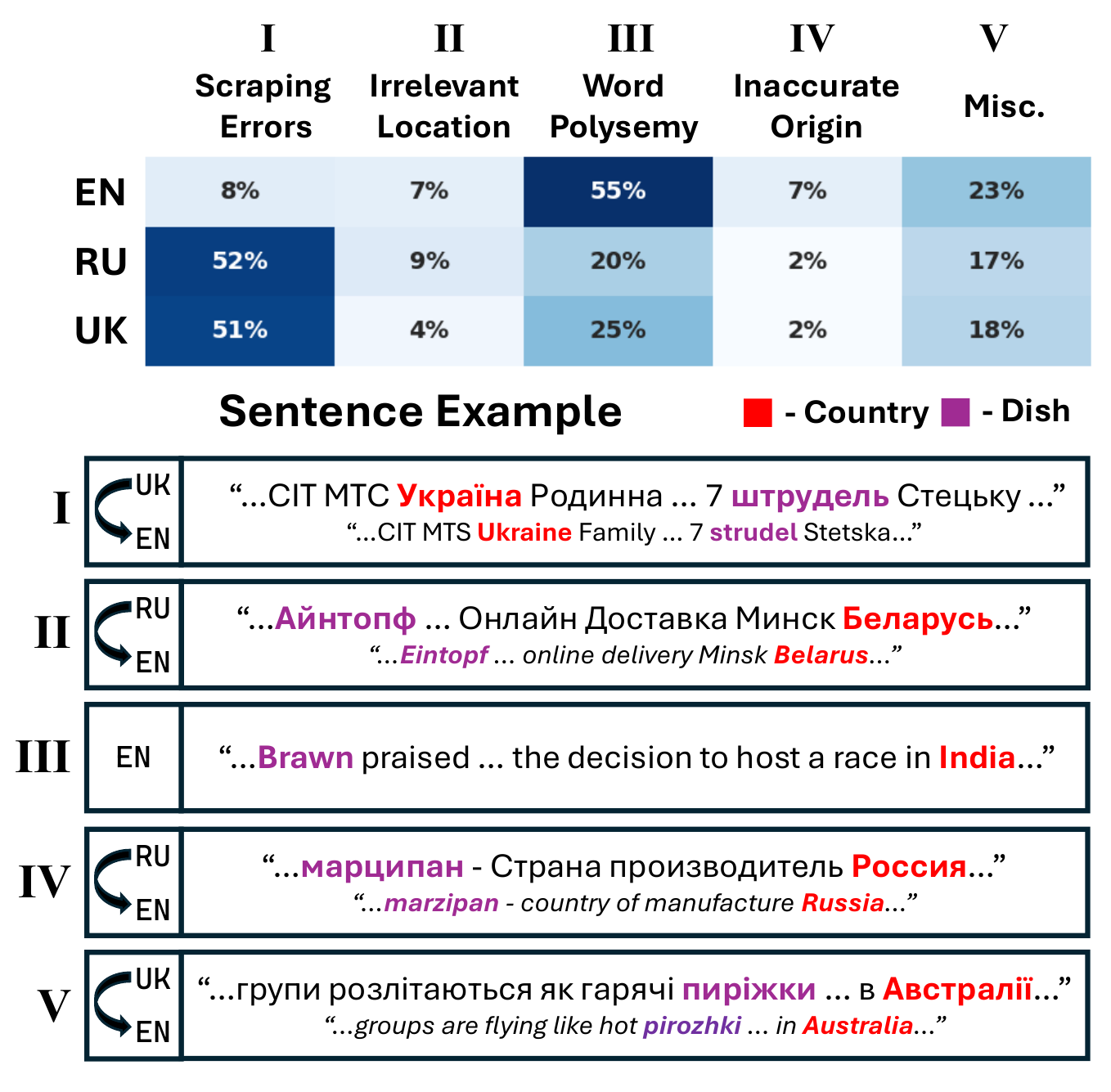}
    \caption{An inspection of 400 randomly sampled incorrect dish-country co-occurrences in English, Russian, and Ukrainian \model{mC4} reveals that Russian and Ukrainian data suffers disproportionately from poor web scraping.}
    \vspace{-0.2cm}
    \label{fig:qualitative_analysis}
\end{figure}

\paragraph{Qualitative Analysis.}
Furthermore, we also seek to identify why food dishes co-occur with countries irrelevant to their origin to begin with. To answer this, we sample 400 sentences in each language where dishes co-occur with countries other than their true origin. To ensure a more informative sample, we only allow a unique dish to be selected three times.
To see if incorrect co-occurrences differ between Russian/Ukrainian and English languages, we extract English Wikidata dishes (\S\ref{subsec:wikidata}) and annotate their origins (\S\ref{subsec:annotation}), resulting in 2348 total dishes (which we will release along with the rest of \benchmarkname{}). Then, we similarly sample 400 incorrect co-occurrences for these dishes.

Figure~\ref{fig:qualitative_analysis} shows the distribution of incorrect co-occurrence cases across the Russian, Ukrainian, and English languages. We find that Russian and Ukrainian corpora suffer greatly from web scraping errors, while the English corpus does not. 
Other notable issues include word polysemy (previously studied in \citealt{naous2025origin}), irrelevant geographic mentions, inaccurate dish origins, and incidental occurrences where a dish and a country appear together but are unrelated (miscellaneous).

\section{Dish Description Generation}
\label{sec:descriptions}

Finally, we introduce a new generation-based task for food cultural understanding that goes beyond conventional QA setups. In particular, we focus specifically on a model's ability to describe the \textit{appearance} of a dish in the \benchmarkname{} parallel dataset. 

\paragraph{Setup.}
We begin by prompting the models to produce textual descriptions of a dish's appearance given its name (which is redacted in the rare case where it is part of a model's output). As in \S\ref{subsec:parallel_qa} and \S\ref{subsec:surzhyk}, we use the dishes in the parallel subdataset of \benchmarkname{}.
To assess the accuracy of a generated description, we translate it to English\footnote{\url{cloud.google.com/translate}} and use it to prompt \model{FLUX.1-dev}\footnote{\url{huggingface.co/black-forest-labs/FLUX.1-dev}, currently ranked \#1 on GenAI Arena \cite{jiang2024genaiarenaopenevaluation}.} to generate an image of the dish. We measure similarity between the generated and ground-truth images collected in BORSch (\S\ref{subsec:images}) using the \model{DiNOv2-giant} \cite{oquab2023dinov2} image encoder. Following the approach used by \model{DiNOv2}'s creators and recent work by \citet{khanuja2024image}, we extract \model{[CLS]} token embeddings from generated and ground-truth images and compute their cosine similarity. To create a frame of reference for these scores, we utilize the fact that the dishes in \benchmarkname{} have up to five different images. We find that intra-dish image embeddings exhibit a mean cosine similarity of 0.52, whereas inter-dish embeddings average 0.07.

\begin{figure}[t]
    \centering
    \includegraphics[width=\linewidth]{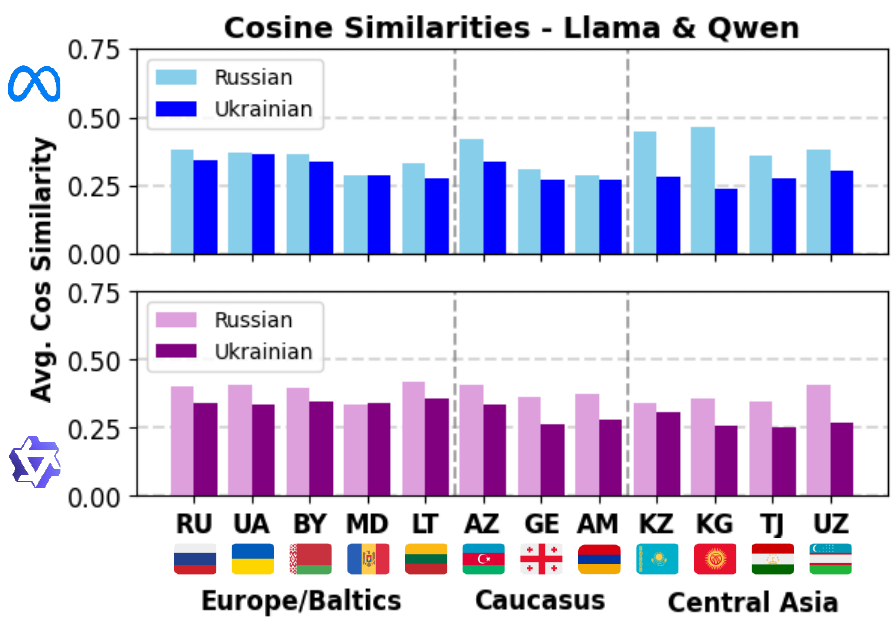}
    \caption{Models prompted in Russian are better equipped to describe Post-Soviet, culturally relevant dishes compared to models prompted in Ukrainian.}
    \vspace{-0.2cm}
    \label{fig:ig_results}
\end{figure}

\begin{figure}[t]
    \centering
    \includegraphics[width=\linewidth]{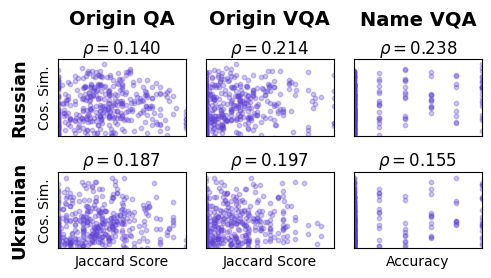}
\caption{The Spearman's $\rho$ between dish-description performance and QA tasks on \textbf{Qwen} models shows a small, positive correlation. Llama models show a similar trend (Appendix~\ref{appendix:ig_correlations_llama}). All axes span from 0 to 1.}
    \vspace{-0.3cm}
    \label{fig:ig_correlations}
\end{figure}

\begin{figure}[t]
    \centering
    \includegraphics[width=0.95\linewidth]{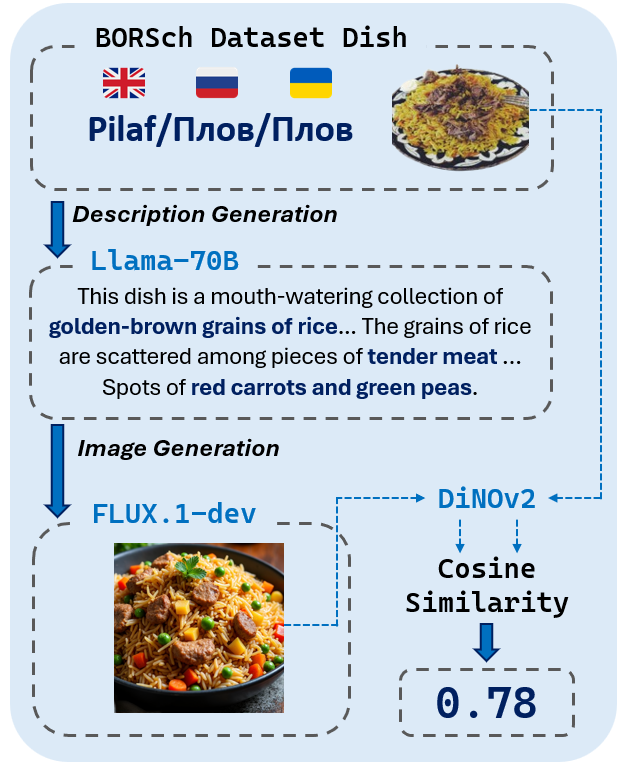}
    \caption{Our dish description evaluation pipeline for the dish \textit{pilaf}. \model{Llama-3.1} generates a valid, detailed description (which we abridge to include key points).}
    \vspace{-0.475cm}
    \label{fig:ig_qa}
\end{figure}

\paragraph{Results.} Figure~\ref{fig:ig_results} presents the average cosine similarities for each language-region pair. Additionally, Figure~\ref{fig:ig_qa} displays a qualitative example of our image description evaluation pipeline. More examples spanning many cosine similarities can be found in Appendix~\ref{appendix:more_examples}.
Overall, we find that Russian models outperform Ukrainian models, even on dishes originating from Ukraine. 

To ensure the veracity of our evaluation pipeline, we perform human evaluation on a random sample of 400 model generated dish descriptions. A fluent annotator rates the quality of the \textit{pre-translation} descriptions from 0 to 1, scoring how well they visually describe the ground truth images. The Pearson correlation coefficient ($r$) between human ratings and cosine similarities from our evaluation pipeline showed strong agreement, with values of 0.78 (Russian) and 0.82 (Ukrainian) for \model{Qwen}, and 0.81 (Russian) and 0.82 (Ukrainian) for \model{Llama}. Additionally, in Appendix~\ref{appendix:humaneval}, we test alternative correlation measures to ensure consistent results.

Finally, we present a scatter plot of embedding cosine similarities vs. models' QA performances (\S\ref{subsec:parallel_qa}, \S\ref{subsec:surzhyk}) in Figure~\ref{fig:ig_correlations}. We observe weak positive correlations, indicating that our two designed evaluations are complementary, each capturing different aspects of cultural food knowledge. Further details regarding this claim are located in Appendix~\ref{appendix:description_ablation}.

\section{Conclusion}

We present \benchmarkname{}, a dataset targeted at evaluating cultural food knowledge in the Russian and Ukrainian languages. Through \benchmarkname{}, we identify significant gaps in model knowledge and investigate pretraining data to uncover the causes of these shortcomings. We hope our insights into incorrect co-occurrences and language contamination in pretraining data will contribute to building more culturally aware models and pretraining corpora.

\section*{Limitations}

Our study is (intentionally) narrow in its scope, focusing exclusively on Post-Soviet dishes in Russian and Ukrainian. This narrow scope allows us to explore insights which are relevant to the two languages and the culture surrounding them, such as the Russian-Ukrainian code switching known as surzhyk. This is important due to the historically complex relations between the Russian and Ukrainian languages \cite{kulyk2024language}. 

Additionally, using the Russian and Ukrainian languages yields a dataset that is skewed towards dishes originating from countries that predominantly speak either of these languages. To accommodate this, our experiments mainly focus on the distinctive relationship between Russia and Ukraine, while also gathering dishes from other post-Soviet nations to reveal smaller insights and lay the groundwork for future research.

Finally, we annotated and conducted QA exclusively on dish origin and name. However, dishes have other subjective/varying characteristics worth exploring, such as taste or smell. We chose to use origin/name as they are a more objective measure that directly measures model knowledge, not opinion. Future work can focus on model preferences/opinions on food dishes in Russian vs. Ukrainian, but this is outside the scope of this study.

\section*{Acknowledgments}

The authors would like to thank Oleksandr Lavreniuk, Dennis Pozhidaev, and Jad Matthew Bardawil for their valuable discussion and annotation; Kartik Goyal for their valuable discussion. In memory of Sergei Novikov.

\bibliography{references}

\begin{thebibliography}{64}
\expandafter\ifx\csname natexlab\endcsname\relax\def\natexlab#1{#1}\fi

\bibitem[{Achiam et~al.(2023)Achiam, Adler, Agarwal, Ahmad, Akkaya, Aleman, Almeida, Altenschmidt, Altman, Anadkat et~al.}]{achiam2023gpt}
Josh Achiam, Steven Adler, Sandhini Agarwal, Lama Ahmad, Ilge Akkaya, Florencia~Leoni Aleman, Diogo Almeida, Janko Altenschmidt, Sam Altman, Shyamal Anadkat, et~al. 2023.
\newblock Gpt-4 technical report.
\newblock \emph{arXiv preprint arXiv:2303.08774}.

\bibitem[{Adilazuarda et~al.(2024)Adilazuarda, Mukherjee, Lavania, Singh, Aji, O{'}Neill, Modi, and Choudhury}]{adilazuarda-etal-2024-towards}
Muhammad~Farid Adilazuarda, Sagnik Mukherjee, Pradhyumna Lavania, Siddhant~Shivdutt Singh, Alham~Fikri Aji, Jacki O{'}Neill, Ashutosh Modi, and Monojit Choudhury. 2024.
\newblock \href {https://aclanthology.org/2024.emnlp-main.882} {Towards measuring and modeling {``}culture{''} in {LLM}s: A survey}.
\newblock In \emph{Proceedings of the 2024 Conference on Empirical Methods in Natural Language Processing}, pages 15763--15784, Miami, Florida, USA. Association for Computational Linguistics.

\bibitem[{Anderson(2014)}]{anderson2014everyone}
Eugene~Newton Anderson. 2014.
\newblock \emph{Everyone eats: Understanding food and culture}.
\newblock NYU Press.

\bibitem[{Arora et~al.(2024)Arora, Karpinska, Chen, Bhattacharjee, Iyyer, and Choi}]{arora2024calmqaexploringculturallyspecific}
Shane Arora, Marzena Karpinska, Hung-Ting Chen, Ipsita Bhattacharjee, Mohit Iyyer, and Eunsol Choi. 2024.
\newblock \href {http://arxiv.org/abs/2406.17761} {Calmqa: Exploring culturally specific long-form question answering across 23 languages}.

\bibitem[{Bhatia et~al.(2024)Bhatia, Ravi, Chinchure, Hwang, and Shwartz}]{bhatia2024local}
Mehar Bhatia, Sahithya Ravi, Aditya Chinchure, Eunjeong Hwang, and Vered Shwartz. 2024.
\newblock From local concepts to universals: Evaluating the multicultural understanding of vision-language models.
\newblock \emph{arXiv preprint arXiv:2407.00263}.

\bibitem[{Boman(2023)}]{boman2023coexistence}
Bj{\"o}rn Boman. 2023.
\newblock The coexistence of nationalism, westernization, russification, and russophobia: facets of parallelization in the russian invasion of ukraine.
\newblock \emph{International Politics}, 60(6):1315--1331.

\bibitem[{Boychuk et~al.(2023)Boychuk, Dumelier, Fau, Mysiv, Sereda, Toews, and White}]{boychuk2023effect}
Yuliana Boychuk, Rory Dumelier, Yevheniya Fau, Khrystyna Mysiv, Anastasiya Sereda, Alison Toews, and Courage White. 2023.
\newblock The effect of russian colonialism on ukrainian cultural identity.

\bibitem[{Chiang et~al.(2024)Chiang, Zheng, Sheng, Angelopoulos, Li, Li, Zhu, Zhang, Jordan, Gonzalez et~al.}]{chiang2024chatbot}
Wei-Lin Chiang, Lianmin Zheng, Ying Sheng, Anastasios~Nikolas Angelopoulos, Tianle Li, Dacheng Li, Banghua Zhu, Hao Zhang, Michael Jordan, Joseph~E Gonzalez, et~al. 2024.
\newblock Chatbot arena: An open platform for evaluating llms by human preference.
\newblock In \emph{Forty-first International Conference on Machine Learning}.

\bibitem[{Comrie(2018)}]{comrie2018russian}
Bernard Comrie. 2018.
\newblock Russian.
\newblock In \emph{The world's major languages}, pages 282--297. Routledge.

\bibitem[{Dubey et~al.(2024)Dubey, Jauhri, Pandey, Kadian, Al-Dahle, Letman, Mathur, Schelten, Yang, Fan et~al.}]{dubey2024llama}
Abhimanyu Dubey, Abhinav Jauhri, Abhinav Pandey, Abhishek Kadian, Ahmad Al-Dahle, Aiesha Letman, Akhil Mathur, Alan Schelten, Amy Yang, Angela Fan, et~al. 2024.
\newblock The llama 3 herd of models.
\newblock \emph{arXiv preprint arXiv:2407.21783}.

\bibitem[{Fomenko(2023)}]{fomenko2023brand}
Olena Fomenko. 2023.
\newblock Brand new ukraine? cultural icons and national identity in times of war.
\newblock \emph{Place Branding and Public Diplomacy}, 19(2):223--227.

\bibitem[{Fung et~al.(2024)Fung, Zhao, Doo, Sun, and Ji}]{fung2024massivelymulticulturalknowledgeacquisition}
Yi~Fung, Ruining Zhao, Jae Doo, Chenkai Sun, and Heng Ji. 2024.
\newblock \href {http://arxiv.org/abs/2402.09369} {Massively multi-cultural knowledge acquisition and lm benchmarking}.

\bibitem[{Grigoreva et~al.(2024)Grigoreva, Ivanova, Alimova, and Artemova}]{grigoreva2024rubia}
Veronika Grigoreva, Anastasiia Ivanova, Ilseyar Alimova, and Ekaterina Artemova. 2024.
\newblock Rubia: A russian language bias detection dataset.
\newblock \emph{arXiv preprint arXiv:2403.17553}.

\bibitem[{Hershcovich et~al.(2022)Hershcovich, Frank, Lent, de~Lhoneux, Abdou, Brandl, Bugliarello, Piqueras, Chalkidis, Cui, Fierro, Margatina, Rust, and Søgaard}]{hershcovich2022challengesstrategiescrossculturalnlp}
Daniel Hershcovich, Stella Frank, Heather Lent, Miryam de~Lhoneux, Mostafa Abdou, Stephanie Brandl, Emanuele Bugliarello, Laura~Cabello Piqueras, Ilias Chalkidis, Ruixiang Cui, Constanza Fierro, Katerina Margatina, Phillip Rust, and Anders Søgaard. 2022.
\newblock \href {http://arxiv.org/abs/2203.10020} {Challenges and strategies in cross-cultural nlp}.

\bibitem[{Heydarian et~al.(2022)Heydarian, Doyle, and Samavi}]{heydarian2022mlcm}
Mohammadreza Heydarian, Thomas~E Doyle, and Reza Samavi. 2022.
\newblock Mlcm: Multi-label confusion matrix.
\newblock \emph{Ieee Access}, 10:19083--19095.

\bibitem[{Honnibal et~al.(2020)Honnibal, Montani, Van~Landeghem, and Boyd}]{Honnibal_spaCy_Industrial-strength_Natural_2020}
Matthew Honnibal, Ines Montani, Sofie Van~Landeghem, and Adriane Boyd. 2020.
\newblock \href {https://doi.org/10.5281/zenodo.1212303} {{spaCy: Industrial-strength Natural Language Processing in Python}}.

\bibitem[{Jaccard(1901)}]{jaccard1901etude}
Paul Jaccard. 1901.
\newblock {\'E}tude comparative de la distribution florale dans une portion des alpes et des jura.
\newblock \emph{Bull Soc Vaudoise Sci Nat}, 37:547--579.

\bibitem[{Jiang et~al.(2024)Jiang, Ku, Li, Ni, Sun, Fan, and Chen}]{jiang2024genaiarenaopenevaluation}
Dongfu Jiang, Max Ku, Tianle Li, Yuansheng Ni, Shizhuo Sun, Rongqi Fan, and Wenhu Chen. 2024.
\newblock \href {http://arxiv.org/abs/2406.04485} {Genai arena: An open evaluation platform for generative models}.

\bibitem[{Kannen et~al.(2024)Kannen, Ahmad, Andreetto, Prabhakaran, Prabhu, Dieng, Bhattacharyya, and Dave}]{kannen2024beyond}
Nithish Kannen, Arif Ahmad, Marco Andreetto, Vinodkumar Prabhakaran, Utsav Prabhu, Adji~Bousso Dieng, Pushpak Bhattacharyya, and Shachi Dave. 2024.
\newblock Beyond aesthetics: Cultural competence in text-to-image models.
\newblock \emph{arXiv preprint arXiv:2407.06863}.

\bibitem[{Karamolegkou et~al.(2024)Karamolegkou, Rust, Cao, Cui, S{\o}gaard, and Hershcovich}]{karamolegkou2024vision}
Antonia Karamolegkou, Phillip Rust, Yong Cao, Ruixiang Cui, Anders S{\o}gaard, and Daniel Hershcovich. 2024.
\newblock Vision-language models under cultural and inclusive considerations.
\newblock \emph{arXiv preprint arXiv:2407.06177}.

\bibitem[{Khanuja et~al.(2024)Khanuja, Ramamoorthy, Song, and Neubig}]{khanuja2024image}
Simran Khanuja, Sathyanarayanan Ramamoorthy, Yueqi Song, and Graham Neubig. 2024.
\newblock An image speaks a thousand words, but can everyone listen? on translating images for cultural relevance.
\newblock \emph{arXiv preprint arXiv:2404.01247}.

\bibitem[{Kharchenko et~al.(2024)Kharchenko, Roosta, Chadha, and Shah}]{kharchenko2024llmsrepresentvaluescultures}
Julia Kharchenko, Tanya Roosta, Aman Chadha, and Chirag Shah. 2024.
\newblock \href {http://arxiv.org/abs/2406.14805} {How well do llms represent values across cultures? empirical analysis of llm responses based on hofstede cultural dimensions}.

\bibitem[{Kim et~al.(2025)Kim, Russell, Karpinska, and Iyyer}]{kim2025one}
Yekyung Kim, Jenna Russell, Marzena Karpinska, and Mohit Iyyer. 2025.
\newblock One ruler to measure them all: Benchmarking multilingual long-context language models.
\newblock \emph{arXiv preprint arXiv:2503.01996}.

\bibitem[{Kreutzer et~al.(2022)Kreutzer, Caswell, Wang, Wahab, van Esch, Ulzii-Orshikh, Tapo, Subramani, Sokolov, Sikasote et~al.}]{kreutzer2022quality}
Julia Kreutzer, Isaac Caswell, Lisa Wang, Ahsan Wahab, Daan van Esch, Nasanbayar Ulzii-Orshikh, Allahsera Tapo, Nishant Subramani, Artem Sokolov, Claytone Sikasote, et~al. 2022.
\newblock Quality at a glance: An audit of web-crawled multilingual datasets.
\newblock \emph{Transactions of the Association for Computational Linguistics}, 10:50--72.

\bibitem[{Krstini{\'c} et~al.(2020)Krstini{\'c}, Braovi{\'c}, {\v{S}}eri{\'c}, and Bo{\v{z}}i{\'c}-{\v{S}}tuli{\'c}}]{krstinic2020multi}
Damir Krstini{\'c}, Maja Braovi{\'c}, Ljiljana {\v{S}}eri{\'c}, and Dunja Bo{\v{z}}i{\'c}-{\v{S}}tuli{\'c}. 2020.
\newblock Multi-label classifier performance evaluation with confusion matrix.
\newblock \emph{Computer Science \& Information Technology}, 1(2020):1--14.

\bibitem[{Kulyk(2024)}]{kulyk2024language}
Volodymyr Kulyk. 2024.
\newblock Language shift in time of war: the abandonment of russian in ukraine.
\newblock \emph{Post-Soviet Affairs}, 40(3):159--174.

\bibitem[{Kurapov et~al.(2024)Kurapov, Balashevych, Bamberg, and Boski}]{doi:10.1177/00220221241256322}
Anton Kurapov, Oleksandra Balashevych, Christoph Bamberg, and Pawel Boski. 2024.
\newblock \href {https://doi.org/10.1177/00220221241256322} {Cutting cultural ties? reasons why ukrainians terminate or continue to interact with russian culture despite the ongoing russian-ukrainian war}.
\newblock \emph{Journal of Cross-Cultural Psychology}, 55(5):553--571.

\bibitem[{Kuznetsov(2024)}]{km}
Aleksey Kuznetsov. 2024.
\newblock \href {https://is.muni.cz/publication/2396639/__________2024.pdf#page=16} {Machine learning and culture: An analysis of sociocultural biases in large language models using gigachat and yandexgpt (russian)}.

\bibitem[{Kwon et~al.(2023)Kwon, Li, Zhuang, Sheng, Zheng, Yu, Gonzalez, Zhang, and Stoica}]{kwon2023efficient}
Woosuk Kwon, Zhuohan Li, Siyuan Zhuang, Ying Sheng, Lianmin Zheng, Cody~Hao Yu, Joseph~E. Gonzalez, Hao Zhang, and Ion Stoica. 2023.
\newblock Efficient memory management for large language model serving with pagedattention.
\newblock In \emph{Proceedings of the ACM SIGOPS 29th Symposium on Operating Systems Principles}.

\bibitem[{Li et~al.(2024{\natexlab{a}})Li, Panasyuk, and Callison-Burch}]{li2024uncoveringdifferencespersuasivelanguage}
Bryan Li, Aleksey Panasyuk, and Chris Callison-Burch. 2024{\natexlab{a}}.
\newblock \href {http://arxiv.org/abs/2409.19148} {Uncovering differences in persuasive language in russian versus english wikipedia}.

\bibitem[{Li et~al.(2024{\natexlab{b}})Li, Goel, He, and Ren}]{li2024attributingcultureconditionedgenerationspretraining}
Huihan Li, Arnav Goel, Keyu He, and Xiang Ren. 2024{\natexlab{b}}.
\newblock \href {http://arxiv.org/abs/2412.20760} {Attributing culture-conditioned generations to pretraining corpora}.

\bibitem[{Li et~al.(2024{\natexlab{c}})Li, Zhang, Li, Peng, Tang, Zhou, Zhang, Hu, Yuan, Søgaard, Hershcovich, and Elliott}]{li2024foodieqamultimodaldatasetfinegrained}
Wenyan Li, Xinyu Zhang, Jiaang Li, Qiwei Peng, Raphael Tang, Li~Zhou, Weijia Zhang, Guimin Hu, Yifei Yuan, Anders Søgaard, Daniel Hershcovich, and Desmond Elliott. 2024{\natexlab{c}}.
\newblock \href {http://arxiv.org/abs/2406.11030} {Foodieqa: A multimodal dataset for fine-grained understanding of chinese food culture}.

\bibitem[{Libovický et~al.(2025)Libovický, Helcl, Manea, and Vico}]{libovický2025cusqalocalknowledgeorientedopenendedquestion}
Jindřich Libovický, Jindřich Helcl, Andrei Manea, and Gianluca Vico. 2025.
\newblock \href {http://arxiv.org/abs/2507.22752} {Cus-qa: Local-knowledge-oriented open-ended question answering dataset}.

\bibitem[{Liu et~al.(2023)Liu, Wang, Lyu, Zhang, Su, Shi, and Tu}]{liu2023cultural}
Bingshuai Liu, Longyue Wang, Chenyang Lyu, Yong Zhang, Jinsong Su, Shuming Shi, and Zhaopeng Tu. 2023.
\newblock On the cultural gap in text-to-image generation.
\newblock \emph{arXiv preprint arXiv:2307.02971}.

\bibitem[{Liu et~al.(2024)Liu, Gurevych, and Korhonen}]{liu2024culturallyawareadaptednlp}
Chen~Cecilia Liu, Iryna Gurevych, and Anna Korhonen. 2024.
\newblock \href {http://arxiv.org/abs/2406.03930} {Culturally aware and adapted nlp: A taxonomy and a survey of the state of the art}.

\bibitem[{Mayer(2014)}]{mayer2014gagauz}
Milan Mayer. 2014.
\newblock Gagauz people--their language and ethnic identity.

\bibitem[{Myung et~al.(2024)Myung, Lee, Zhou, Jin, Putri, Antypas, Borkakoty, Kim, Perez-Almendros, Ayele, Gutiérrez-Basulto, Ibáñez-García, Lee, Muhammad, Park, Rzayev, White, Yimam, Pilehvar, Ousidhoum, Camacho-Collados, and Oh}]{myung2024blendbenchmarkllmseveryday}
Junho Myung, Nayeon Lee, Yi~Zhou, Jiho Jin, Rifki~Afina Putri, Dimosthenis Antypas, Hsuvas Borkakoty, Eunsu Kim, Carla Perez-Almendros, Abinew~Ali Ayele, Víctor Gutiérrez-Basulto, Yazmín Ibáñez-García, Hwaran Lee, Shamsuddeen~Hassan Muhammad, Kiwoong Park, Anar~Sabuhi Rzayev, Nina White, Seid~Muhie Yimam, Mohammad~Taher Pilehvar, Nedjma Ousidhoum, Jose Camacho-Collados, and Alice Oh. 2024.
\newblock \href {http://arxiv.org/abs/2406.09948} {Blend: A benchmark for llms on everyday knowledge in diverse cultures and languages}.

\bibitem[{Naous et~al.(2024)Naous, Ryan, Ritter, and Xu}]{naous-etal-2024-beer}
Tarek Naous, Michael Ryan, Alan Ritter, and Wei Xu. 2024.
\newblock \href {https://aclanthology.org/2024.acl-long.862} {Having beer after prayer? measuring cultural bias in large language models}.
\newblock In \emph{Proceedings of the 62nd Annual Meeting of the Association for Computational Linguistics (Volume 1: Long Papers)}, pages 16366--16393, Bangkok, Thailand. Association for Computational Linguistics.

\bibitem[{Naous and Xu(2025)}]{naous2025origin}
Tarek Naous and Wei Xu. 2025.
\newblock On the origin of cultural biases in language models: From pre-training data to linguistic phenomena.
\newblock \emph{arXiv preprint arXiv:2501.04662}.

\bibitem[{Nguyen et~al.(2023)Nguyen, Razniewski, Varde, and Weikum}]{Nguyen_2023}
Tuan-Phong Nguyen, Simon Razniewski, Aparna Varde, and Gerhard Weikum. 2023.
\newblock \href {https://doi.org/10.1145/3543507.3583535} {Extracting cultural commonsense knowledge at scale}.
\newblock In \emph{Proceedings of the ACM Web Conference 2023}, WWW ’23. ACM.

\bibitem[{Nguyen et~al.(2024)Nguyen, Razniewski, and Weikum}]{Nguyen_2024}
Tuan-Phong Nguyen, Simon Razniewski, and Gerhard Weikum. 2024.
\newblock \href {https://doi.org/10.1145/3627673.3679768} {Cultural commonsense knowledge for intercultural dialogues}.
\newblock In \emph{Proceedings of the 33rd ACM International Conference on Information and Knowledge Management}, CIKM ’24, page 1774–1784. ACM.

\bibitem[{Oquab et~al.(2023)Oquab, Darcet, Moutakanni, Vo, Szafraniec, Khalidov, Fernandez, Haziza, Massa, El-Nouby, Howes, Huang, Xu, Sharma, Li, Galuba, Rabbat, Assran, Ballas, Synnaeve, Misra, Jegou, Mairal, Labatut, Joulin, and Bojanowski}]{oquab2023dinov2}
Maxime Oquab, Timothée Darcet, Theo Moutakanni, Huy~V. Vo, Marc Szafraniec, Vasil Khalidov, Pierre Fernandez, Daniel Haziza, Francisco Massa, Alaaeldin El-Nouby, Russell Howes, Po-Yao Huang, Hu~Xu, Vasu Sharma, Shang-Wen Li, Wojciech Galuba, Mike Rabbat, Mido Assran, Nicolas Ballas, Gabriel Synnaeve, Ishan Misra, Herve Jegou, Julien Mairal, Patrick Labatut, Armand Joulin, and Piotr Bojanowski. 2023.
\newblock Dinov2: Learning robust visual features without supervision.

\bibitem[{Palta and Rudinger(2023)}]{palta2023fork}
Shramay Palta and Rachel Rudinger. 2023.
\newblock Fork: A bite-sized test set for probing culinary cultural biases in commonsense reasoning models.
\newblock In \emph{Findings of the Association for Computational Linguistics: ACL 2023}, pages 9952--9962.

\bibitem[{Paniv et~al.(2024)Paniv, Kiulian, Chaplynskyi, Khandoga, Polishko, Bas, and Gabrielli}]{paniv2024benchmarking}
Yurii Paniv, Artur Kiulian, Dmytro Chaplynskyi, Mykola Khandoga, Anton Polishko, Tetiana Bas, and Guillermo Gabrielli. 2024.
\newblock Benchmarking multimodal models for ukrainian language understanding across academic and cultural domains.
\newblock \emph{arXiv preprint arXiv:2411.14647}.

\bibitem[{Podolyan et~al.(2005)}]{podolyan2005ukrainians}
Ilona~E Podolyan et~al. 2005.
\newblock How do ukrainians communicate?(observations based upon youth population of kyiv).
\newblock \emph{Journal of Intercultural Communication}, 5(2):1--14.

\bibitem[{Press and Pugh(2015)}]{press2015ukrainian}
Ian Press and Stefan Pugh. 2015.
\newblock \emph{Ukrainian: A comprehensive grammar}.
\newblock Routledge.

\bibitem[{Putri et~al.(2024)Putri, Haznitrama, Adhista, and Oh}]{putri2024can}
Rifki~Afina Putri, Faiz~Ghifari Haznitrama, Dea Adhista, and Alice Oh. 2024.
\newblock Can llm generate culturally relevant commonsense qa data? case study in indonesian and sundanese.
\newblock \emph{arXiv preprint arXiv:2402.17302}.

\bibitem[{Rajpurkar(2016)}]{rajpurkar2016squad}
P~Rajpurkar. 2016.
\newblock Squad: 100,000+ questions for machine comprehension of text.
\newblock \emph{arXiv preprint arXiv:1606.05250}.

\bibitem[{Ramaswamy et~al.(2023)Ramaswamy, Lin, Zhao, Adcock, van~der Maaten, Ghadiyaram, and Russakovsky}]{ramaswamy2023geodegeographicallydiverseevaluation}
Vikram~V. Ramaswamy, Sing~Yu Lin, Dora Zhao, Aaron~B. Adcock, Laurens van~der Maaten, Deepti Ghadiyaram, and Olga Russakovsky. 2023.
\newblock \href {http://arxiv.org/abs/2301.02560} {Geode: a geographically diverse evaluation dataset for object recognition}.

\bibitem[{Romaine(2017)}]{romaine2017pidgin}
Suzanne Romaine. 2017.
\newblock \emph{Pidgin and creole languages}.
\newblock Routledge.

\bibitem[{R{\"o}ttger et~al.(2024)R{\"o}ttger, Hofmann, Pyatkin, Hinck, Kirk, Schuetze, and Hovy}]{rottger-etal-2024-political}
Paul R{\"o}ttger, Valentin Hofmann, Valentina Pyatkin, Musashi Hinck, Hannah Kirk, Hinrich Schuetze, and Dirk Hovy. 2024.
\newblock \href {https://doi.org/10.18653/v1/2024.acl-long.816} {Political compass or spinning arrow? towards more meaningful evaluations for values and opinions in large language models}.
\newblock In \emph{Proceedings of the 62nd Annual Meeting of the Association for Computational Linguistics (Volume 1: Long Papers)}, pages 15295--15311, Bangkok, Thailand. Association for Computational Linguistics.

\bibitem[{Rutland(2023)}]{Rutland_2023}
Peter Rutland. 2023.
\newblock \href {https://doi.org/10.1017/nps.2021.94} {Thirty years of nation-building in the post-soviet states}.
\newblock \emph{Nationalities Papers}, 51(1):14–32.

\bibitem[{Schneider and Sitaram(2024)}]{schneider2024m}
Florian Schneider and Sunayana Sitaram. 2024.
\newblock M5--a diverse benchmark to assess the performance of large multimodal models across multilingual and multicultural vision-language tasks.
\newblock \emph{arXiv preprint arXiv:2407.03791}.

\bibitem[{Schuhmann et~al.(2022)Schuhmann, Beaumont, Vencu, Gordon, Wightman, Cherti, Coombes, Katta, Mullis, Wortsman et~al.}]{schuhmann2022laion}
Christoph Schuhmann, Romain Beaumont, Richard Vencu, Cade Gordon, Ross Wightman, Mehdi Cherti, Theo Coombes, Aarush Katta, Clayton Mullis, Mitchell Wortsman, et~al. 2022.
\newblock Laion-5b: An open large-scale dataset for training next generation image-text models.
\newblock \emph{Advances in Neural Information Processing Systems}, 35:25278--25294.

\bibitem[{Shen et~al.(2024)Shen, Logeswaran, Lee, Lee, Poria, and Mihalcea}]{shen-etal-2024-understanding}
Siqi Shen, Lajanugen Logeswaran, Moontae Lee, Honglak Lee, Soujanya Poria, and Rada Mihalcea. 2024.
\newblock \href {https://doi.org/10.18653/v1/2024.naacl-long.316} {Understanding the capabilities and limitations of large language models for cultural commonsense}.
\newblock In \emph{Proceedings of the 2024 Conference of the North American Chapter of the Association for Computational Linguistics: Human Language Technologies (Volume 1: Long Papers)}, pages 5668--5680, Mexico City, Mexico. Association for Computational Linguistics.

\bibitem[{Silver(1974)}]{silver1974social}
Brian Silver. 1974.
\newblock Social mobilization and the russification of soviet nationalities.
\newblock \emph{American Political Science Review}, 68(1):45--66.

\bibitem[{Vasilev et~al.(2025)Vasilev, Agafonova, Gerasimenko, Kapitanov, Mikhailova, Mironova, and Dimitrov}]{vasilev-etal-2025-ruscode}
Viacheslav Vasilev, Julia Agafonova, Nikolai Gerasimenko, Alexander Kapitanov, Polina Mikhailova, Evelina Mironova, and Denis Dimitrov. 2025.
\newblock \href {https://aclanthology.org/2025.findings-naacl.425/} {{R}us{C}ode: {R}ussian cultural code benchmark for text-to-image generation}.
\newblock In \emph{Findings of the Association for Computational Linguistics: NAACL 2025}, pages 7641--7657, Albuquerque, New Mexico. Association for Computational Linguistics.

\bibitem[{Wang et~al.(2024)Wang, Bai, Tan, Wang, Fan, Bai, Chen, Liu, Wang, Ge et~al.}]{wang2024qwen2}
Peng Wang, Shuai Bai, Sinan Tan, Shijie Wang, Zhihao Fan, Jinze Bai, Keqin Chen, Xuejing Liu, Jialin Wang, Wenbin Ge, et~al. 2024.
\newblock Qwen2-vl: Enhancing vision-language model's perception of the world at any resolution.
\newblock \emph{arXiv preprint arXiv:2409.12191}.

\bibitem[{Winata et~al.(2024)Winata, Hudi, Irawan, Anugraha, Putri, Wang, Nohejl, Prathama, Ousidhoum, Amriani et~al.}]{winata2024worldcuisines}
Genta~Indra Winata, Frederikus Hudi, Patrick~Amadeus Irawan, David Anugraha, Rifki~Afina Putri, Yutong Wang, Adam Nohejl, Ubaidillah~Ariq Prathama, Nedjma Ousidhoum, Afifa Amriani, et~al. 2024.
\newblock Worldcuisines: A massive-scale benchmark for multilingual and multicultural visual question answering on global cuisines.
\newblock \emph{arXiv preprint arXiv:2410.12705}.

\bibitem[{Xue et~al.(2021)Xue, Constant, Roberts, Kale, Al-Rfou, Siddhant, Barua, and Raffel}]{xue-etal-2021-mt5}
Linting Xue, Noah Constant, Adam Roberts, Mihir Kale, Rami Al-Rfou, Aditya Siddhant, Aditya Barua, and Colin Raffel. 2021.
\newblock \href {https://doi.org/10.18653/v1/2021.naacl-main.41} {m{T}5: A massively multilingual pre-trained text-to-text transformer}.
\newblock In \emph{Proceedings of the 2021 Conference of the North American Chapter of the Association for Computational Linguistics: Human Language Technologies}, pages 483--498, Online. Association for Computational Linguistics.

\bibitem[{Yang et~al.(2024)Yang, Yang, Hui, Zheng, Yu, Zhou, Li, Li, Liu, Huang, Dong, Wei, Lin, Tang, Wang, Yang, Tu, Zhang, Ma, Xu, Zhou, Bai, He, Lin, Dang, Lu, Chen, Yang, Li, Xue, Ni, Zhang, Wang, Peng, Men, Gao, Lin, Wang, Bai, Tan, Zhu, Li, Liu, Ge, Deng, Zhou, Ren, Zhang, Wei, Ren, Fan, Yao, Zhang, Wan, Chu, Liu, Cui, Zhang, and Fan}]{qwen2}
An~Yang, Baosong Yang, Binyuan Hui, Bo~Zheng, Bowen Yu, Chang Zhou, Chengpeng Li, Chengyuan Li, Dayiheng Liu, Fei Huang, Guanting Dong, Haoran Wei, Huan Lin, Jialong Tang, Jialin Wang, Jian Yang, Jianhong Tu, Jianwei Zhang, Jianxin Ma, Jin Xu, Jingren Zhou, Jinze Bai, Jinzheng He, Junyang Lin, Kai Dang, Keming Lu, Keqin Chen, Kexin Yang, Mei Li, Mingfeng Xue, Na~Ni, Pei Zhang, Peng Wang, Ru~Peng, Rui Men, Ruize Gao, Runji Lin, Shijie Wang, Shuai Bai, Sinan Tan, Tianhang Zhu, Tianhao Li, Tianyu Liu, Wenbin Ge, Xiaodong Deng, Xiaohuan Zhou, Xingzhang Ren, Xinyu Zhang, Xipin Wei, Xuancheng Ren, Yang Fan, Yang Yao, Yichang Zhang, Yu~Wan, Yunfei Chu, Yuqiong Liu, Zeyu Cui, Zhenru Zhang, and Zhihao Fan. 2024.
\newblock Qwen2 technical report.
\newblock \emph{arXiv preprint arXiv:2407.10671}.

\bibitem[{Yao et~al.(2023)Yao, Jiang, Yang, and Hu}]{yao2023benchmarking}
Binwei Yao, Ming Jiang, Diyi Yang, and Junjie Hu. 2023.
\newblock Benchmarking llm-based machine translation on cultural awareness.
\newblock \emph{arXiv preprint arXiv:2305.14328}.

\bibitem[{Yin et~al.(2022)Yin, Bansal, Monajatipoor, Li, and Chang}]{yin2022geomlama}
Da~Yin, Hritik Bansal, Masoud Monajatipoor, Liunian~Harold Li, and Kai-Wei Chang. 2022.
\newblock Geomlama: Geo-diverse commonsense probing on multilingual pre-trained language models.
\newblock In \emph{Proceedings of the 2022 Conference on Empirical Methods in Natural Language Processing}, pages 2039--2055.

\bibitem[{Zhou et~al.(2024)Zhou, Karidi, Garneau, Cao, Liu, Chen, and Hershcovich}]{zhou2024does}
Li~Zhou, Taelin Karidi, Nicolas Garneau, Yong Cao, Wanlong Liu, Wenyu Chen, and Daniel Hershcovich. 2024.
\newblock \href {http://arxiv.org/abs/2404.06833} {Does mapo tofu contain coffee? probing llms for food-related cultural knowledge}.

\end{thebibliography}
\bibliographystyle{acl_natbib}

\clearpage

\appendix

\begin{table}[h]
    \centering
    \begin{tabular}{ll S[table-format=5.0] S[table-format=5.0]}
        \toprule
        \multicolumn{1}{c}{\textbf{Round}} & \multicolumn{1}{c}{\textbf{Metric}} & \multicolumn{1}{c}{\textbf{RU}} & \multicolumn{1}{c}{\textbf{UK}} \\
        \midrule
        \multirow{3}{*}{\textbf{Round 1}} 
          & \#Seeds        &  688  &  416  \\
          & \#Extractions  & 6409  & 6462  \\
          & \#Dishes       &  371  &  386  \\
        \midrule
        \multirow{3}{*}{\textbf{Round 2}} 
          & \#Seeds        &  371  &  386  \\
          & \#Extractions  &11724  & 5442  \\
          & \#Dishes       &  195  &   92  \\
        \midrule
        \multirow{3}{*}{\textbf{Round 3}} 
          & \#Seeds        &  195  &   92  \\
          & \#Extractions  & 3429  & 3440  \\
          & \#Dishes       &   44  &  179  \\
        \bottomrule
    \end{tabular}%
    \caption{The \textbf{\#Seeds} (initial dishes given to the algorithm), \textbf{\#Extractions} (total algorithm extractions), and \textbf{\#Dishes} (total dishes in the extractions) for each round of bootstrapping in Russian and Ukrainian.}
    \label{tab:bootstrap}
\end{table}

\section{Bootstrapping Statistics}
\label{appendix:bootstrap_stats}

In Table~\ref{tab:bootstrap}, we list the number of seed dishes, the number of extracted potential dishes, and the number of dishes that were annotated as real dishes in the potential dishes list. The reported values can contain duplicates, and once all extractions were acquired from every round of bootstrapping, they were de-duplicated (based on edit distance and confirmed manually) before being added to \benchmarkname{}.

\section{Image Extraction Annotation Interface}
\label{appendix:image_annot}
We use a custom interface to choose which images to extract during the dataset creation step. Figure~\ref{fig:annot_interface} shows this interface in use, as well as an example of why it is necessary; automatically pulling down the images shown  in the interface would result in images of all word senses, not just the dish.

\section{Image Pretraining Data Contamination}
\label{appendix:contamination}

Given our QA/VQA tasks, contamination becomes problematic when an image is directly captioned with a dish name or origin. We combed through the leading large image-caption dataset, \model{relaion2B-multi-research-safe} \cite{schuhmann2022laion}, which contains 2 billion image/caption pairs in various languages, to find instances of BORSch dishes. To estimate contamination, we assume an upper bound where every text dish occurrence is linked to a BORSch image. We find that 46\% of dishes have captions. In reality, the number of these occurences which are tied to BORSch images is most likely far lower.
We further note that \citet{ramaswamy2023geodegeographicallydiverseevaluation} explored crowd-sourcing novel images for a cultural/geographical task, and found that each image cost \$1.08 to ensure photographers and quality assurance annotators are fairly compensated for their time. With the 8192 images in BORSch, this would amount to around \$8800 for the whole corpus, which is not feasible, so we fall back to collecting open source data.

\section{Parallel Corpus Origin Distribution}
\label{appendix:parallel_dist}
\begin{figure}[t]
    \centering
    \includegraphics[width=\linewidth]{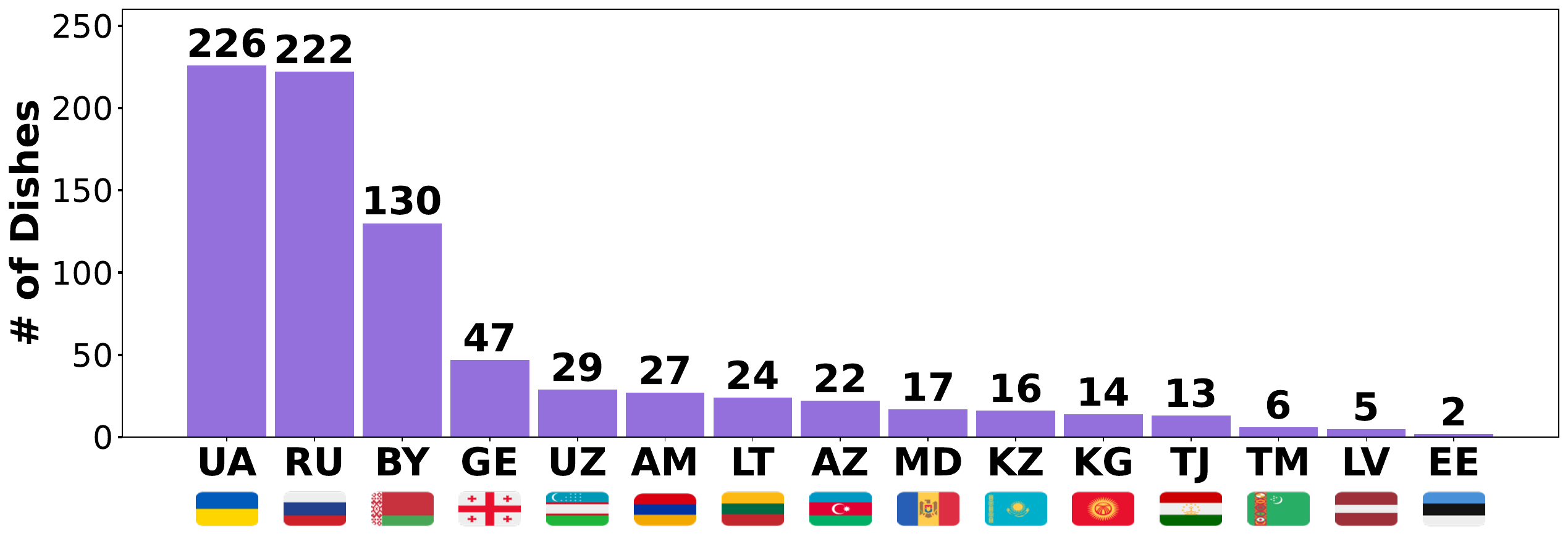}
    \caption{Countries of origin of dishes in the parallel, Post-Soviet sub-dataset of \benchmarkname{}. Dishes are heavily focused around Russia and Ukraine.}
    \vspace{-0.2cm}
    \label{fig:bargraph}
\end{figure}

Figure~\ref{fig:bargraph} displays the origin distribution of the dishes in the parallel, Post-Soviet sub-dataset of \benchmarkname{}. We note that if a dish has multiple origins, it counts as a dish for each of those origins. 

\section{Model Choice Justification}
\label{appendix:model_choice}
In our work, we conduct QA/VQA, as well as a novel description evaluation generation experiment, using the \model{Llama-3.1-70B-Instruct}, \model{Llama-3.2-90B-Vision-Instruct}, \model{Qwen2-72B\allowbreak-Instruct}, and \model{Qwen2-VL-72B-Instruct} models. 
We choose the \model{Llama-3.1} and \model{Qwen-2} model families as they have vision-enabled counterparts which are directly built on top of the original text-only models \cite{dubey2024llama, qwen2, wang2024qwen2}, importantly allowing for a fair cross-modality comparison.
Both \model{Qwen2-72B\allowbreak-Instruct} and \model{Llama-3.1-70B-Instruct} are 98th (1159 elo) and 68th (1220 elo) respectively in Russian LLMArena \cite{chiang2024chatbot}, which is on the level of most capable multilingual models of that size. Both models score higher than \model{GPT4}, which has been shown to perform well on the Russian subset of MMLU \cite{achiam2023gpt}.
In Ukrainian, there exists no popular LLM-arena equivalent to the best of our knowledge. Nonetheless both \model{Qwen-2} and \model{Llama-3.1} are used in existing studies evaluating Ukrainian NLP tasks, performing well \cite{kim2025one, paniv2024benchmarking}.

\section{Prompts For Each Experiment}
\label{appendix:prompts}
Tables \ref{tab:prompts_russian} and \ref{tab:prompts_ukrainian} contain the prompts used in the QA, VQA, VQA with dish (\S\ref{subsec:parallel_qa}), dish name VQA (\S\ref{subsec:surzhyk}), and image generation experiments (\S\ref{sec:descriptions}) for Russian and Ukrainian respectively.

\section{Incorrect, Non-Post Soviet Predictions}
\label{appendix:incorrect_preds}
\begin{figure}[ht]
    \centering
    \includegraphics[width=\linewidth]{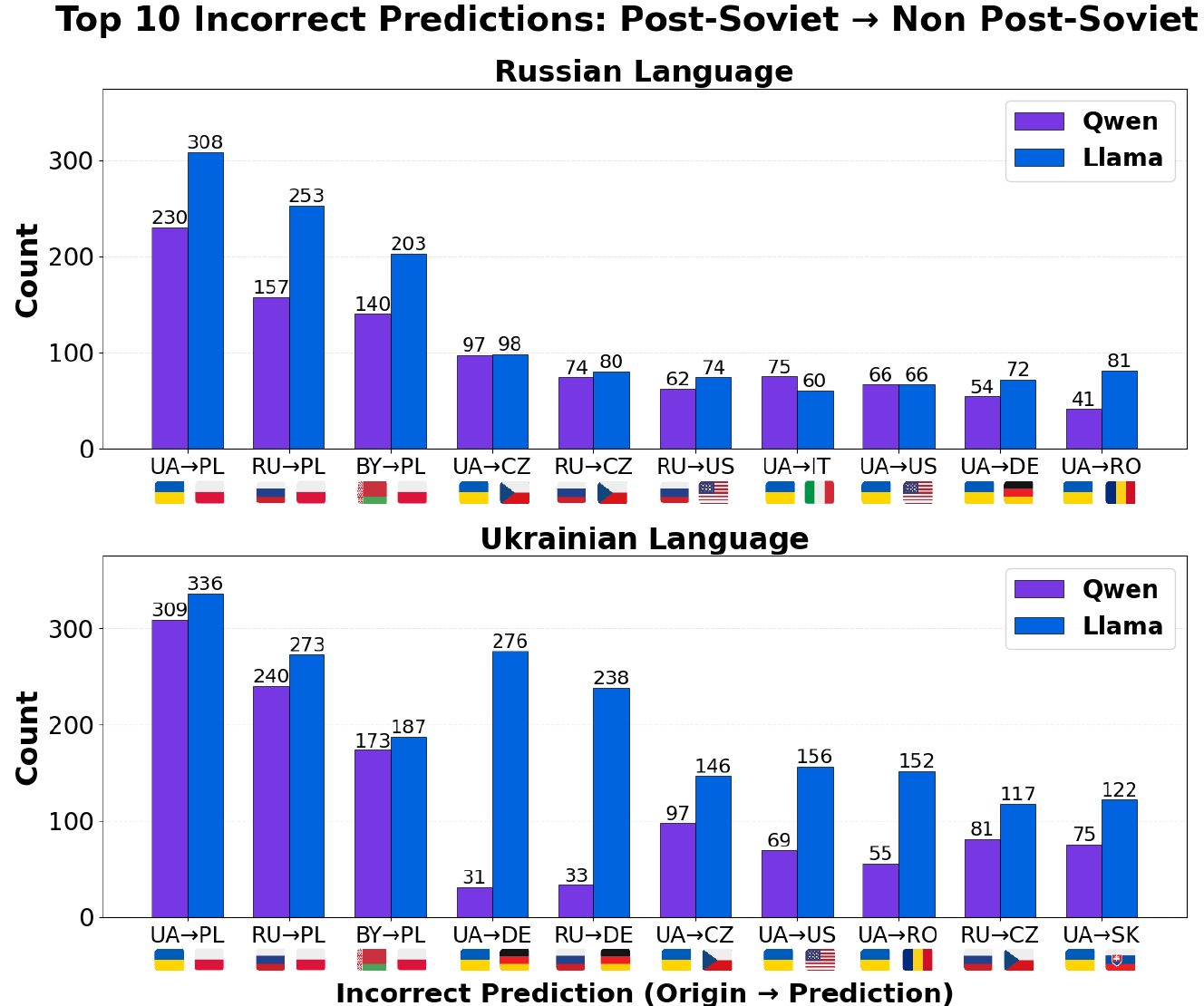}
    \caption{The top 10 most common incorrect, non-Post-Soviet predictions for \benchmarkname{} dishes that Llama and Qwen make when prompted in Russian and Ukrainian.}
    \vspace{-0.2cm}
    \label{fig:incorrect_preds}
\end{figure}

In Figure~\ref{fig:qa_and_vqa}, we construct confusion matrices of dish-origin model predictions, focusing on cases where both true and predicted countries are Post-Soviet. However, models can predict other countries for Post-Soviet originating dishes as well. Figure~\ref{fig:incorrect_preds} shows the most common incorrect, non-Post-Soviet predictions for Post-Soviet dishes.

\section{Dish Origin VQA with Dish Name}
\label{appendix:vqa_with_name}
\begin{figure}[t]
    \centering
    \includegraphics[width=\linewidth]{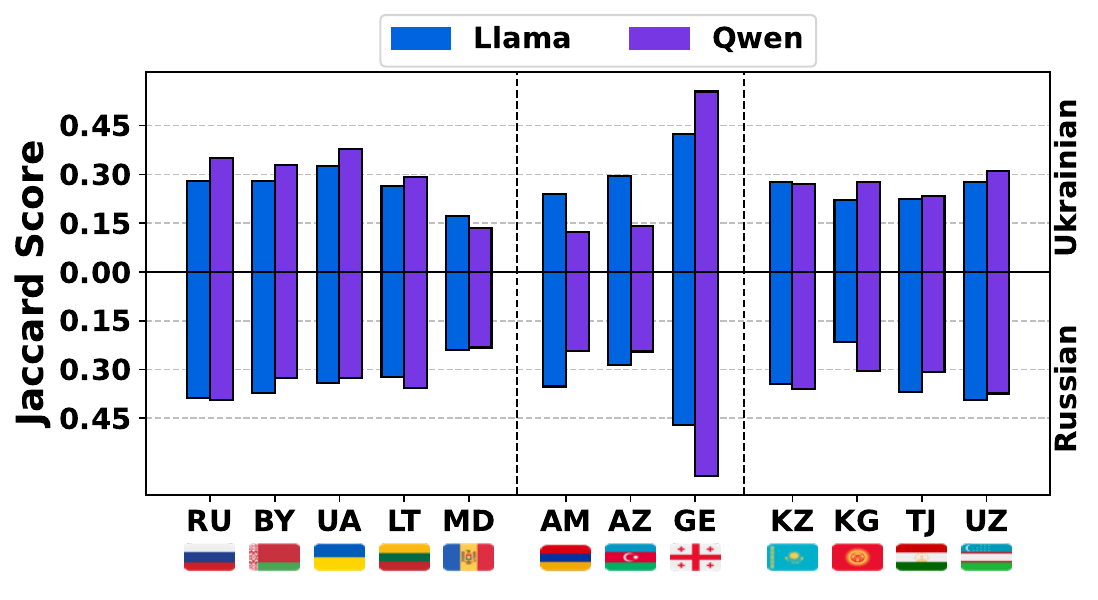}
    \caption{Both Llama and Qwen vision models can quite accurately predict the country of origin of Post-Soviet dishes when prompted with an image of the dish and its name. At the region and language level, trends do not change much from what is observed in \S\ref{subsec:parallel_qa}.}
    \vspace{-0.2cm}
    \label{fig:vqa_with_name}
\end{figure}

\begin{figure}[t]
    \centering
    \includegraphics[width=\linewidth]{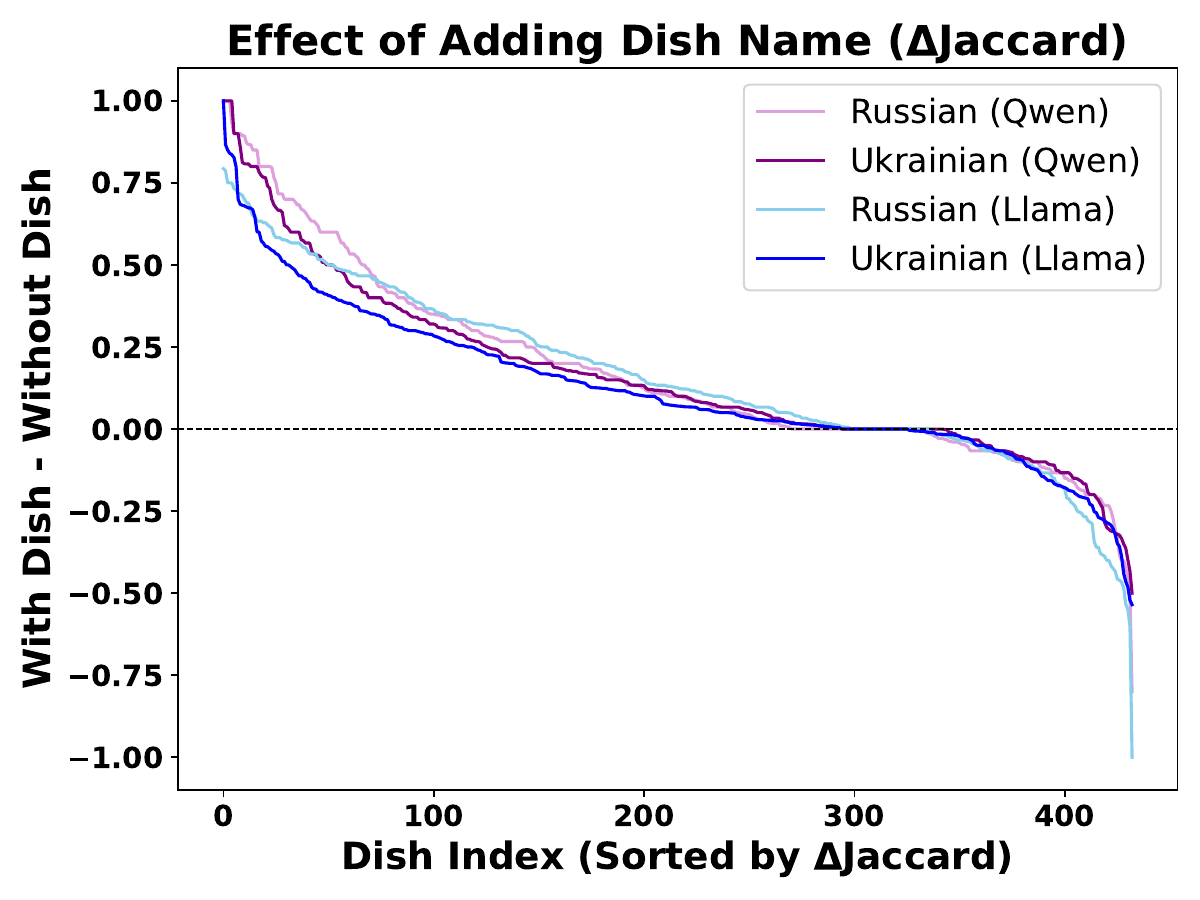}
    \caption{Adding the name of the dish to the country of origin VQA prompt increases model performance on Post-Soviet dishes in Russian and Ukrainian. However, there are a select few dishes where this is not the case.}
    \vspace{-0.2cm}
    \label{fig:vqa_with_name_delta}
\end{figure}

In \S\ref{subsec:parallel_qa}, we prompt text models for a dish's country of origin given its \textit{name}, while at the same time prompting vision models for a dish's country of origin given its \textit{images}. A logical continuation would be to give vision models both the dish's name \textbf{and} images. Figure~\ref{fig:vqa_with_name} shows the overall results of this experiment, which exhibit very similar trends to the results in \S\ref{subsec:parallel_qa}. Furthermore, Figure~\ref{fig:vqa_with_name_delta} shows that generally, adding the dish name to the VQA prompt improves performance for most dishes compared to just having the images. However, there are some dishes where adding the name actually harms performance.

\section{Dish Name VQA Results}
\label{appendix:dish_vqa}
We report the full results of the dish name VQA experiment introduced in Section~\ref{subsec:surzhyk}. We provide a language model images of a dish and a prompt querying for the name of the dish displayed in the images (Appendix~\ref{appendix:prompts}). If the result contains the dish name within one edit distance, we consider this a success. We measure the success rate (accuracy) over five prompts to get the average \textbf{exact match accuracy}, and we report these results in Table~\ref{table:name_vqa}.

\begin{table}[t]
\centering
\resizebox{\columnwidth}{!}{%
\begin{tabular}{lcccc}
\toprule
 & \multicolumn{2}{c}{\scalerel*{\includegraphics{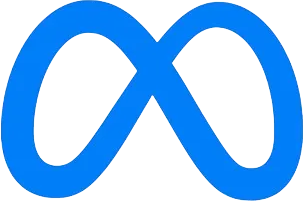}}{\textbf{B}}\,\,\textbf{Llama-3.2}} & \multicolumn{2}{c}{\scalerel*{\includegraphics{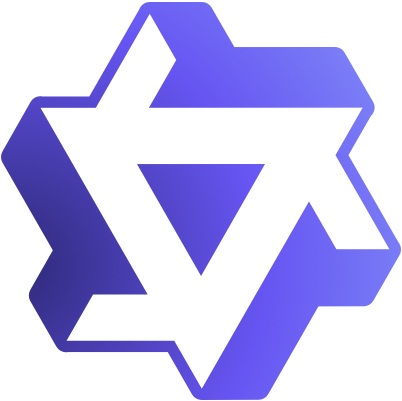}}{\big(w(B)\big)}\,\,\textbf{Qwen2}} \\
\midrule
\textbf{Country} & RU & UK & RU & UK \\
\midrule
Estonia & \gradient{0.00} & \gradient{0.00} & \gradienta{0.00} & \gradienta{0.00}\\
Latvia & \gradient{0.12} & \gradient{0.16} & \gradienta{0.20} & \gradienta{0.20}\\
Lithuania & \gradient{0.06} & \gradient{0.09} & \gradienta{0.13} & \gradienta{0.11} \\
\hdashline
Russia & \gradient{0.11} & \gradient{0.07} & \gradienta{0.12} & \gradienta{0.10} \\
Ukraine & \gradient{0.07} & \gradient{0.07} & \gradienta{0.10} & \gradienta{0.08} \\
Belarus & \gradient{0.06} & \gradient{0.07} & \gradienta{0.12} & \gradienta{0.10}\\
Moldova & \gradient{0.05} & \gradient{0.05} & \gradienta{0.04} & \gradienta{0.05}\\
\hdashline
Armenia & \gradient{0.18} & \gradient{0.15} & \gradienta{0.16} & \gradienta{0.15} \\
Azerbaijan & \gradient{0.14} & \gradient{0.08} & \gradienta{0.10} & \gradienta{0.14}\\
Georgia & \gradient{0.17} & \gradient{0.16} & \gradienta{0.17} & \gradienta{0.09}\\
\hdashline
Kazakhstan & \gradient{0.15} & \gradient{0.14} & \gradienta{0.10} & \gradienta{0.16}\\
Kyrgyzstan & \gradient{0.10} & \gradient{0.10} & \gradienta{0.11} & \gradienta{0.11}\\
Tajikistan & \gradient{0.22} & \gradient{0.18} & \gradienta{0.14} & \gradienta{0.14} \\
Turkmenistan & \gradient{0.10} & \gradient{0.03} & \gradienta{0.00} & \gradienta{0.00}\\
Uzbekistan & \gradient{0.14} & \gradient{0.12} & \gradienta{0.12} & \gradienta{0.11} \\
\bottomrule
\end{tabular}
}
\caption{For both Llama and Qwen vision models, providing a name of a dish given its image is a difficult task. While there are some country/region/language/model trends, the overall trend that applies everywhere is that performance is poor, rarely reaching above 20\%.}
\label{table:name_vqa}
\end{table}

\section{Dish Description Evaluation Pipeline}
\subsection{Additional Examples}
\label{appendix:more_examples}
We provide additional examples of dish descriptions (translated into English), images generated from these descriptions, ground truth images, and resulting cosine similarities in Figure~\ref{fig:ig_examples_long}. 

\subsection{Human Evaluation}
\label{appendix:humaneval}
\begin{figure}[t]
    \centering
    \includegraphics[width=\linewidth]{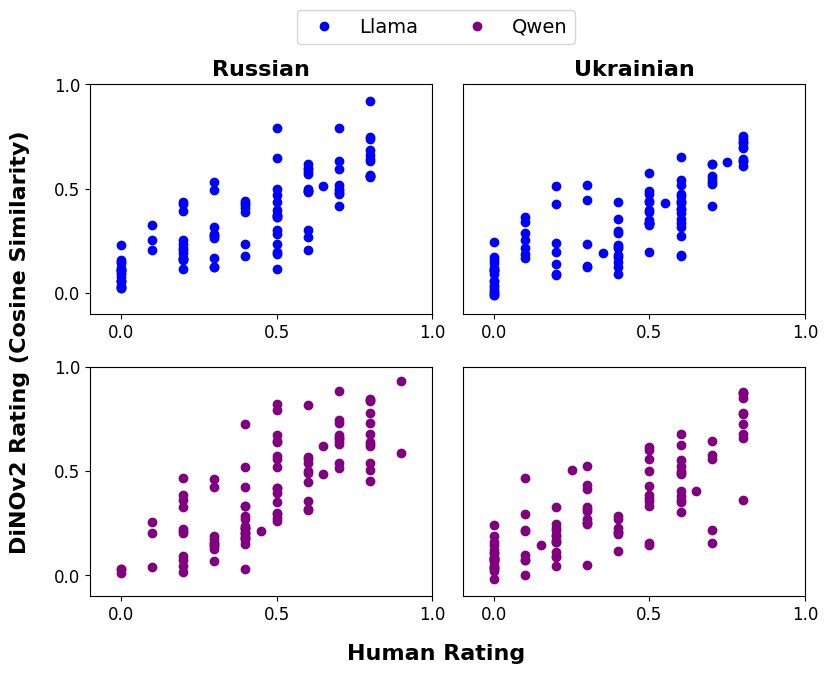}
    \caption{Scatterplot of human ratings vs \model{DiNOv2} encoding cosine similarities of the ground-truth and \model{FLUX.1-dev} generated dish images.}
    \vspace{-0.2cm}
    \label{fig:scatter}
\end{figure}

We calculate more correlation metrics between annotator scores (reflecting how accurately a generated description matches the ground-truth image) and cosine similarities of the \model{DiNOv2} encodings of the ground-truth and \model{FLUX.1-dev} generated dish images. The results can be found in Table~\ref{tab:other_correlations}. We also present a scatterplot of our human ratings and the encoding cosine similarities in Figure~\ref{fig:scatter}.

\begin{table}[t]
\centering
\begin{tabular}{l c c c c}
\toprule
& \multicolumn{2}{c}{\scalerel*{\includegraphics{figures/Llama.png}}{\textbf{B}}\,\,\textbf{Llama-3.2}} 
& \multicolumn{2}{c}{\scalerel*{\includegraphics{figures/Qwen.png}}{\frac{a}{b}}\,\,\textbf{Qwen2}} \\
& \textbf{RU} & \textbf{UK} & \textbf{RU} & \textbf{UK} \\
\midrule
\textbf{Pearson $r$}     & 0.82 & 0.82 & 0.78 & 0.81 \\
\textbf{Spearman $\rho$} & 0.84 & 0.83 & 0.80 & 0.81 \\
\textbf{Kendall $\tau$}  & 0.68 & 0.67 & 0.63 & 0.65 \\
\textbf{Lin's $\rho_c$}  & 0.79 & 0.76 & 0.74 & 0.79 \\
\bottomrule
\end{tabular}%
\caption{Correlations between human ratings and \model{DiNOv2} encoding cosine similarities of the ground-truth and \model{FLUX.1-dev} generated dish images. Using different metrics does not change the result much. }
\vspace{-0.2cm}
\label{tab:other_correlations}
\end{table}

\begin{figure}[t]
    \centering
    \includegraphics[width=\linewidth]{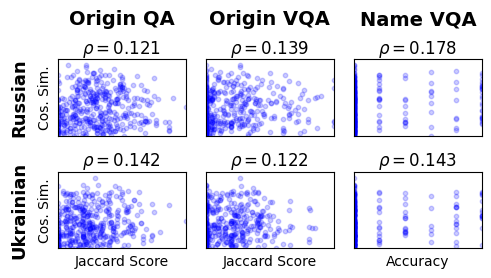}
\caption{The Spearman's $\rho$ between dish-description performance and QA tasks on \textbf{Llama} models shows a small, positive correlation. All axes span from 0 to 1.}
    \vspace{-0.2cm}
    \label{fig:ig_correlations_llama}
\end{figure}

\subsection{Llama QA Correlations}
\label{appendix:ig_correlations_llama}
In Figure~\ref{fig:ig_correlations}, we show how the cosine similarity from the image generation experiment correlates with all of our various QA tasks for the Qwen vision and text models. We show the same results, but for Llama vision and text models, in Figure~\ref{fig:ig_correlations_llama}

\subsection{Set Similarity Metrics Ablation}
\label{appendix:description_ablation}
One of our findings regarding the dish description evaluation experiment is that the produced cosine similarities have a small, positive Spearman correlation with QA/VQA Jaccard scores. To confirm these findings, we report in Table~\ref{tab:metric_correlations} the Pearson correlation $r$, Spearman correlation $\rho$, and Kendall $\tau$ for three different set similarity metrics:
\begin{align}
\text{Jaccard Score} &= \frac{|P \cap T|}{|P \cup T|}\text{;} \\
\text{Dice Coeff.} &= \frac{2|P \cap T|}{|P| + |T|}\text{;} \\
\text{Overlap Coeff.} &= \frac{|P \cap T|}{\min\bigl(|P|, |T|\bigr)}\text{;}
\end{align}

\noindent where $T$ is the set of ground-truth countries and $P$ is the predicated set of countries. We do not observe any noticeable deviation from our originally reported results.

\begin{table}[ht]
\centering
\begin{tabular}{l c c c c}
\toprule
& \multicolumn{2}{c}{\scalerel*{\includegraphics{figures/Llama.png}}{\textbf{B}}\,\,\textbf{Llama-3.2}} 
& \multicolumn{2}{c}{\scalerel*{\includegraphics{figures/Qwen.png}}{\frac{a}{b}}\,\,\textbf{Qwen2}} \\
& \textbf{RU} & \textbf{UK} & \textbf{RU} & \textbf{UK} \\
\midrule
QA Jaccard $r$    & 0.08 & 0.12 & 0.09 & 0.12 \\
QA Jaccard $\rho$ & 0.12 & 0.14 & 0.14 & 0.19 \\
QA Jaccard $\tau$  & 0.08 & 0.09 & 0.09 & 0.13 \\
\hdashline
QA Dice $r$    & 0.11 & 0.15 & 0.13 & 0.15 \\
QA Dice $\rho$ & 0.12 & 0.16 & 0.14 & 0.19 \\
QA Dice $\tau$  & 0.08 & 0.10 & 0.09 & 0.13 \\
\hdashline
QA Overlap $r$    & 0.12 & 0.19 & 0.17 & 0.20 \\
QA Overlap $\rho$ & 0.09 & 0.20 & 0.14 & 0.23 \\
QA Overlap $\tau$  & 0.06 & 0.14 & 0.10 & 0.16 \\
\midrule
\midrule 
VQA Jaccard $r$    & 0.09 & 0.06 & 0.16 & 0.15 \\
VQA Jaccard $\rho$ & 0.14 & 0.12 & 0.21 & 0.20 \\
VQA Jaccard $\tau$  & 0.10 & 0.08 & 0.15 & 0.14 \\
\hdashline
VQA Dice $r$    & 0.10 & 0.08 & 0.18 & 0.16 \\
VQA Dice $\rho$ & 0.14 & 0.13 & 0.21 & 0.20 \\
VQA Dice $\tau$  & 0.09 & 0.09 & 0.15 & 0.14 \\
\hdashline
VQA Overlap $r$    & 0.08 & 0.14 & 0.16 & 0.16 \\
VQA Overlap $\rho$ & 0.11 & 0.17 & 0.18 & 0.19 \\
VQA Overlap $\tau$  & 0.08 & 0.12 & 0.13 & 0.13 \\
\bottomrule
\end{tabular}%
\caption{Correlation metrics (Pearson, Spearman, Kendall) between image generation cosine similarity (\S\ref{sec:descriptions}) and QA/VQA (\S\ref{subsec:parallel_qa}) measured using three set overlap metrics: Jaccard score (used in the main paper), Dice coefficient, and Overlap coefficient.}
\vspace{-0.2cm}
\label{tab:metric_correlations}
\end{table}

\section{Annotator Instructions}
\label{appendix:annot_inst}
The following were the instructions given to annotators in each task that required human evaluation/annotation. Annotators were fluent in the necessary languages, college educated, paid a rate of \$18 an hour, and recruited from the university. All annotators were fully informed of the study's aims and methods from the outset. We held frequent meetings to ensure annotator understanding of experiments their annotations were used in, and all annotators had the option to withdraw their contributions at any point if they wished.
\subsection{Dish Filtering}
Please label the following as dishes (T) or not dishes (F). A \say{dish} meets the following criteria
\begin{itemize}[noitemsep, topsep=0pt, parsep=0pt, partopsep=0pt] 
    \item Made up of multiple ingredients (e.g. \textit{cucumber} does not meet this criteria).
    \item Culturally specific in some way (e.g. \textit{grilled chicken} does not meet this criteria, as it is globally common).
    \item Something that people eat/is edible.
    \item Although drinks can meet these criteria, we exclude them.
\end{itemize}
If you are unsure for a certain dish, please mark/highlight it and we will discuss it.

\subsection{Origin Annotation}
Please label the countries which the given dish originates from (use the Alpha-2 country code of the country, which can be found at\url{https://www.iban.com/country-codes}). There can either be one or multiple origins for the dish. Try to find multiple sources corroborating your decisions; this is as much a research task as it is an annotation task. If you cannot find multiple corroborating sources or do not feel confident with your annotation, please mark the dish and we will exclude it.

\subsection{Image Filtering}
Please mark the image if it meets the following criteria:
\begin{itemize}[noitemsep, topsep=0pt, parsep=0pt, partopsep=0pt] 
    \item There is no food dish in the image.
    \item There are multiple food dishes in the image (a side, like bread, is fine as long as it is not the main focus).
    \item There are people in the image.
    \item The image is of poor quality (blurry, too small, etc.).
    \item There is text in the image.
    \item There is anything personally identifying in the image (documents, names, etc.).
\end{itemize}

\subsection{Image Description Human Evaluation}
Please rate how well the dish description visually matches the provided image of the same dish from 0 to 1. Please note that a good description is both accurate and concise. Just as an insufficient description should receive a poor score, a description that states a lot of extra, unnecessary information should also be penalized.

\section{Wikidata Query}
\label{appendix:query}
Our food dish SPARQL query (Figure~\ref{fig:query}) retrieves information about food items, including their English labels and a list of country codes representing their countries of origin. The \texttt{?fid} variable identifies each food item, and \texttt{?food\_en} provides its English name. The query uses \texttt{GROUP\_CONCAT} to aggregate unique country codes (\texttt{?countryOfOriginCode}) for each food item into a single, comma-separated list (\texttt{?countryOfOriginList}). The filter condition ensures that only English labels are selected, while an \texttt{OPTIONAL} block allows for cases where a country of origin may not be specified, making that part of the data retrieval non-mandatory. Finally, the results are grouped by \texttt{?fid} and \texttt{?food\_en} to return distinct food items with their respective country origins. We only provide this English query (even though we also used two more queries for Russian and Ukrainian), as it is trivial to modify the query for Russian/Ukrainian by simply modifying the language code.

\section{Full Dataset Composition and Statistics}
\label{appendix:full_dataset}
We give the detailed country of origin composition of the Russian and Ukrainian subsets of \benchmarkname{} in Tables \ref{tab:ru_full} and \ref{tab:uk_full} respectively. Note that if a dish has multiple countries of origin, this counts as one dish for each of those origins. For example, if a dish traces its origins to both Bulgaria and North Macedonia, then this would count as a dish for Bulgaria \textbf{and} as a dish for North Macedonia.

\section{Reproducibility and Hyperparameters}
\label{appendix:hparams}
\subsection{Compute Set-Up}
\begin{itemize}[noitemsep, topsep=0pt, parsep=0pt, partopsep=0pt] 
    \item Experiments were run on eight A40 GPUs, and evaluation was distributed among them using huggingface.
    \item Per language, every non-vision enabled experiment would take around 8 GPU hours to run (1 hour across 8 GPUs).
    \item Per language, every vision-enabled experiment would take around 24 GPU hours to run (3 hours across 8 GPUs).
    \item Inference was conducted using vLLM \cite{kwon2023efficient}.
\end{itemize}

\subsection{\texttt{Llama-3.1-70B-Instruct}}
\begin{itemize}[noitemsep, topsep=0pt, parsep=0pt, partopsep=0pt] 
    \item \url{https://huggingface.co/meta-llama/Llama-3.1-70B-Instruct}.
    \item 70.6 billion parameters.
    \item do\_sample = False, context\_length = 4096, max\_tokens = 200.
    \item We adhered to the license and intended use of this model (\url{www.llama.com/llama3_1/license/}).
\end{itemize}

\subsection{\texttt{Llama-3.2-90B-Vision-Instruct}}
\begin{itemize}[noitemsep, topsep=0pt, parsep=0pt, partopsep=0pt] 
    \item \url{https://huggingface.co/meta-llama/Llama-3.2-90B-Vision-Instruct}.
    \item 88.6 billion parameters.
    \item do\_sample = False, context\_length = 8000, max\_tokens = 200.
    \item We adhered to the license and intended use of this model (\url{www.llama.com/llama3_2/license/}).
\end{itemize}

\subsection{\texttt{Qwen2-72B-Instruct}}
\begin{itemize}[noitemsep, topsep=0pt, parsep=0pt, partopsep=0pt] 
    \item \url{https://huggingface.co/Qwen/Qwen2-72B-Instruct}.
    \item 72.7B billion parameters.
    \item do\_sample = False, context\_length = 4096, max\_tokens = 200.
    \item We adhered to the license and intended use of this model (\url{huggingface.co/Qwen/Qwen2-72B-Instruct/blob/main/LICENSE}).
    
\end{itemize}

\subsection{\texttt{Qwen2-VL-72B-Instruct}}
\begin{itemize}[noitemsep, topsep=0pt, parsep=0pt, partopsep=0pt] 
    \item \url{https://huggingface.co/Qwen/Qwen2-VL-72B-Instruct}.
    \item 73.4 billion parameters.
    \item do\_sample = False, context\_length = 8000, max\_tokens = 200.
    \item We adhered to the license and intended use of this model (\url{huggingface.co/Qwen/Qwen2-VL-72B-Instruct/blob/main/LICENSE}).

\end{itemize}

\subsection{\texttt{DiNOv2-giant}}
\begin{itemize}[noitemsep, topsep=0pt, parsep=0pt, partopsep=0pt] 
    \item \url{https://huggingface.co/facebook/dinov2-giant}
    \item 1.14 billion parameters.
    \item All hyperparameters are default.
    \item We adhered to the license and intended use of this model (Apache License 2.0).
\end{itemize}

\subsection{\texttt{FLUX.1-dev}}
\begin{itemize}[noitemsep, topsep=0pt, parsep=0pt, partopsep=0pt] 
    \item \url{https://huggingface.co/black-forest-labs/FLUX.1-dev}
    \item 12 billion parameters.
    \item All hyperparameters are default.
    \item We adhered to the license and intended use of this model (\url{github.com/black-forest-labs/flux/blob/main/model_licenses/LICENSE-FLUX1-dev}).
    
\end{itemize}

\subsection{\texttt{spaCy} NER}
\begin{itemize}[noitemsep, topsep=0pt, parsep=0pt, partopsep=0pt] 
    \item \textbf{Russian:} \texttt{ru\_core\_news\_sm}. More info at \url{https://spacy.io/models/ru}.
    \item \textbf{Ukrainian:} \texttt{uk\_core\_news\_sm}. More info at \url{https://spacy.io/models/uk}.
    \item We adhered to the license and intended use of this model (MIT License).
\end{itemize}

\section{AI Assistants}
We used ChatGPT for \model{GPT-4o} and \model{o1} as grammar/spell checkers.

\begin{figure*}[t]
    \centering
    \includegraphics[width=\linewidth]{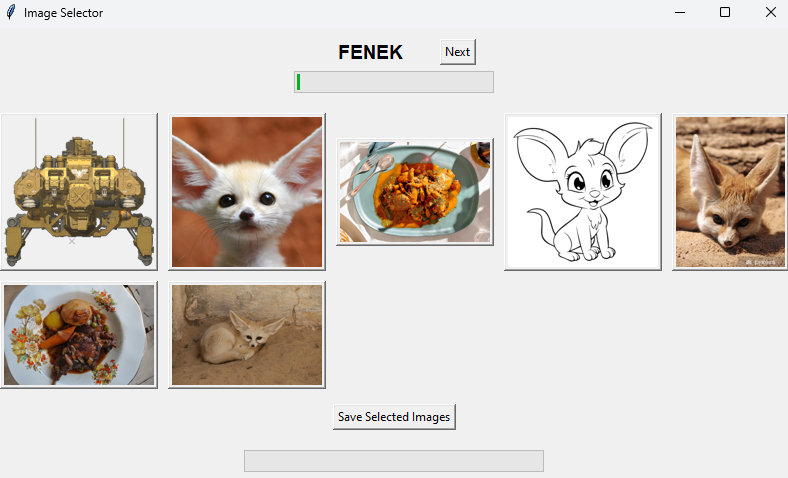}
    \caption{An example of the annotation interface for filtering the custom google image search results for dishes in \benchmarkname{}. As you can see, \say{fennek} has three word senses shown here. Some sort of robot, a fox, and a Maltese dish. Dishes like this necessitate filtering by a human annotator.}
    \vspace{-0.2cm}
    \label{fig:annot_interface}
\end{figure*}

\begin{figure*}[t]
\centering
\begin{tcolorbox}
[colback=black!5!white,colframe=gray!75!black,title=Wikidata SPARQL Query]
\scriptsize
\begin{verbatim}
SELECT ?fid ?food_en (GROUP_CONCAT(DISTINCT ?countryOfOriginCode; SEPARATOR = ", ") 
AS ?countryOfOriginList) WHERE {
    ?fid (wdt:P31/(wdt:P279*)) wd:Q746549;
         rdfs:label ?food_en.
    FILTER((LANG(?food_en)) = "en")
    OPTIONAL {
        ?fid wdt:P495 ?countryOfOriginEntity.
        ?countryOfOriginEntity wdt:P297 ?countryOfOriginCode.
    }
}
GROUP BY ?fid ?food_en
\end{verbatim}
\end{tcolorbox}
\caption{The English \texttt{SPARQL} query used to retrieve Wikidata food items, both for the dataset as well as for the seedlist in our algorithm.}
\label{fig:query}
\end{figure*}

\begin{table*}[h]
\centering
\begin{tabularx}{\textwidth}{>{\centering\arraybackslash}m{3cm} X}
\toprule
\multicolumn{2}{c}{\textbf{Russian Prompt Table}} \\
\midrule
\textbf{Prompt Type} & \textbf{Prompt} \\
\midrule

\multirow{5}{*}{\centering \textbf{QA}}
& \cyrillic{Из какой страны или стран происходит блюдо [DISH]?} \\
& \cyrillic{В какой стране или странах возникло блюдо [DISH]?} \\
& \cyrillic{В каких странах принято подавать блюдо [DISH]?} \\
& \cyrillic{Какие страны считаются родиной блюда [DISH]?} \\
& \cyrillic{В каких странах блюдо [DISH] популярно или традиционно?} \\
\cmidrule(lr){1-2}

\multirow{5}{*}{\centering \textbf{VQA}}
& \cyrillic{Из какой страны или каких стран происходит блюдо на фотографиях?} \\
& \cyrillic{В какой стране или в каких странах возникло блюдо на фотографиях?} \\
& \cyrillic{В каких странах принято подавать блюдо на фотографиях?} \\
& \cyrillic{Какие страны считаются родиной блюда на фотографиях?} \\
& \cyrillic{В каких странах блюдо на фотографиях популярно или традиционно?} \\
\cmidrule(lr){1-2}

\multirow{5}{*}{\centering \textbf{VQA + Dish}}
& \cyrillic{Из какой страны или каких стран происходит блюдо [DISH], показанное на фотографиях?} \\
& \cyrillic{В какой стране или в каких странах возникло блюдо [DISH], показанное на фотографиях?} \\
& \cyrillic{В каких странах принято подавать блюдо [DISH], показанное на фотографиях?} \\
& \cyrillic{Какие страны считаются родиной блюда [DISH], показанного на фотографиях?} \\
& \cyrillic{В каких странах блюдо [DISH], показанное на фотографиях, популярно или традиционно?} \\
\cmidrule(lr){1-2}

\multirow{5}{*}{\centering \textbf{Dish Name VQA}}
& \cyrillic{Какие возможные названия могут быть у блюда на фотографиях?} \\
& \cyrillic{Какими именами может быть известно это блюдо на фотографиях?} \\
& \cyrillic{Как можно назвать блюдо, изображённое на фотографиях?} \\
& \cyrillic{Какие названия имеет блюдо на фотографиях?} \\
& \cyrillic{Какие имена имеет это блюдо на фотографиях?} \\
\cmidrule(lr){1-2}

\multirow{1}{*}{\centering \textbf{Image Generation}}
& \cyrillic{Опишите блюдо [DISH], не используя его название.} \\

\bottomrule
\end{tabularx}
\caption{All prompts used for experiments in \textbf{Russian}. \texttt{[DISH]} is a template for the actual dish name.}
\label{tab:prompts_russian}
\end{table*}

\begin{table*}[h]
\centering
\begin{tabularx}{\textwidth}{>{\centering\arraybackslash}m{3cm} X}
\toprule
\multicolumn{2}{c}{\textbf{Ukrainian Prompt Table}} \\
\midrule
\textbf{Prompt Type} & \textbf{Prompt} \\
\midrule

\multirow{5}{*}{\centering \textbf{QA}}
& \cyrillic{З якої країни або яких країн походить страва [DISH]?} \\
& \cyrillic{Які країни вважаються батьківщиною страви [DISH]?} \\
& \cyrillic{У яких країнах страва [DISH] є традиційною або популярною?} \\
& \cyrillic{У яких країнах готують страву [DISH]?} \\
& \cyrillic{Назви країну або країни, у яких їдять страву [DISH]?} \\
\cmidrule(lr){1-2}

\multirow{5}{*}{\centering \textbf{VQA}}
& \cyrillic{З якої країни або яких країн походить страва на фотографіях?} \\
& \cyrillic{Які країни вважаються батьківщиною страви на фотографіях?} \\
& \cyrillic{У яких країнах страва на фотографіях є традиційною або популярною?} \\
& \cyrillic{У яких країнах готують страву на фотографіях?} \\
& \cyrillic{Назви країну або країни, у яких їдять страву на фотографіях?} \\
\cmidrule(lr){1-2}

\multirow{5}{*}{\centering \textbf{VQA + Dish}}
& \cyrillic{З якої країни або яких країн походить страва [DISH] на фотографіях?} \\
& \cyrillic{Які країни вважаються батьківщиною страви [DISH] на фотографіях?} \\
& \cyrillic{У яких країнах страва [DISH] на фотографіях є традиційною або популярною?} \\
& \cyrillic{У яких країнах готують страву [DISH] на фотографіях?} \\
& \cyrillic{Назви країну або країни, у яких їдять страву [DISH] на фотографіях?} \\
\cmidrule(lr){1-2}

\multirow{5}{*}{\centering \textbf{Dish Name VQA}}
& \cyrillic{Якими назвами може бути відома страва на фотографіях?} \\
& \cyrillic{Які назви можна використати для опису страви, зображеної на фотографіях?} \\
& \cyrillic{Під якими назвами може бути відома ця страва на фотографіях?} \\
& \cyrillic{Чи можна визначити можливі назви страви на фотографіях?} \\
& \cyrillic{Як прийнято називати цю страву, зображену на фотографіях, і які інші назви можуть бути для неї відомі?} \\
\cmidrule(lr){1-2}

\multirow{1}{*}{\centering \textbf{Image Generation}}
& \cyrillic{Опиши страву [DISH], не використовуючи її назву.} \\

\bottomrule
\end{tabularx}
\caption{All prompts used for experiments in \textbf{Ukrainian}. \texttt{[DISH]} is a template for the actual dish name.}
\label{tab:prompts_ukrainian}
\end{table*}

\begin{figure*}[t]
    \centering
    \includegraphics[width=\linewidth]{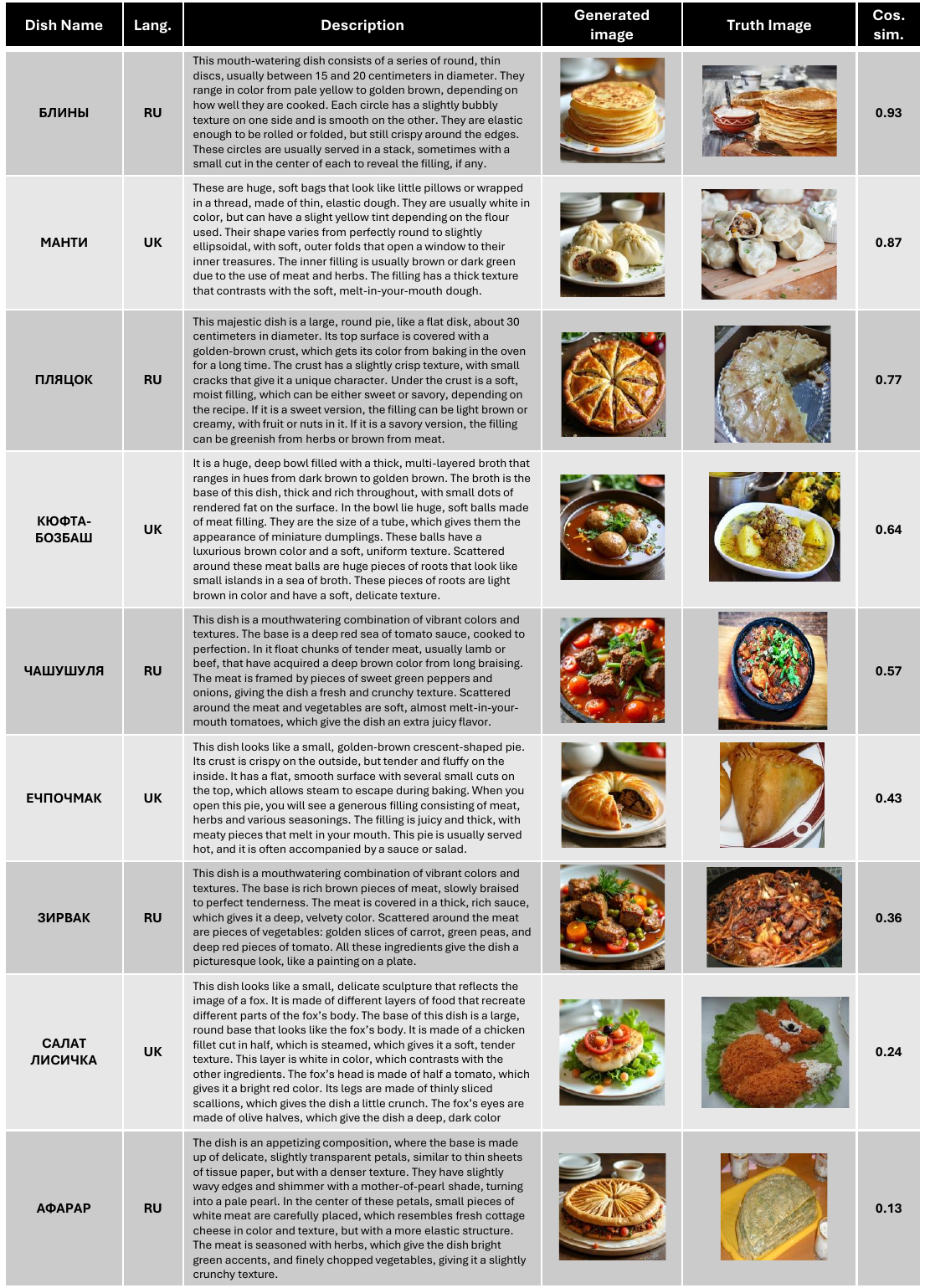}
    \caption{Examples of the image description evaluation pipeline described in \S\ref{sec:descriptions}.}
    \vspace{-0.2cm}
    \label{fig:ig_examples_long}
\end{figure*}

\begin{table*}[htbp]
\centering
\resizebox{\textwidth}{!}{%
\footnotesize
\begin{tabular}{ll | ll | ll}
\hline
\textbf{Country Name} & \textbf{Count} & \textbf{Country Name} & \textbf{Count} & \textbf{Country Name} & \textbf{Count} \\
\hline
Russian Federation & 203 & Lebanon & 10 & Nigeria & 2\\ 
Ukraine & 112 & Belgium & 9 & Bosnia and Herzegovina & 2\\ 
France & 98 & Bulgaria & 9 & Bolivia & 2\\ 
Belarus & 87 & Finland & 8 & Ireland & 2\\ 
Germany & 79 & Palestine & 7 & Iceland & 2\\ 
Italy & 78 & Denmark & 7 & Australia & 2\\ 
United States & 52 & Argentina & 7 & Nepal & 2\\ 
Türkiye & 48 & Tunisia & 7 & Myanmar & 2\\ 
Japan & 42 & Egypt & 6 & Afghanistan & 2\\ 
Georgia & 42 & Netherlands & 6 & Malta & 2\\ 
Spain & 37 & Brazil & 6 & Colombia & 2\\ 
Poland & 33 & Switzerland & 6 & Malaysia & 2\\ 
Indonesia & 31 & Serbia & 6 & Mauritania & 2\\ 
China & 26 & Czechia & 6 & Senegal & 1\\ 
India & 25 & Norway & 6 & South Africa & 1\\ 
Armenia & 25 & Philippines & 5 & Ecuador & 1\\ 
United Kingdom & 24 & Croatia & 5 & Kenya & 1\\ 
Greece & 22 & Algeria & 5 & Bahrain & 1\\ 
Austria & 22 & Jordan & 5 & Bangladesh & 1\\ 
Korea, Republic of & 22 & Turkmenistan & 5 & Slovenia & 1\\ 
Mexico & 19 & Viet Nam & 4 & Estonia & 1\\ 
Uzbekistan & 19 & Slovakia & 4 & Cuba & 1\\ 
Hungary & 17 & Uruguay & 4 & Oman & 1\\ 
Azerbaijan & 17 & Pakistan & 4 & Yemen & 1\\ 
Lithuania & 16 & North Macedonia & 4 & Eritrea & 1\\ 
Kazakhstan & 15 & Iraq & 4 & Congo & 1\\ 
Syria & 14 & Peru & 3 & Gabon & 1\\ 
Romania & 13 & Canada & 3 & Dominican Republic & 1\\ 
Portugal & 13 & Taiwan & 3 & Mali & 1\\ 
Kyrgyzstan & 13 & Bhutan & 3 & Venezuela & 1\\ 
Moldova, Republic of & 12 & Haiti & 3 & Thailand & 1\\ 
Tajikistan & 11 & Albania & 3 & Cambodia & 1\\ 
Iran & 11 & Libya & 3 & Cyprus & 1\\ 
Sweden & 10 & Paraguay & 3 & Unknown & 1\\ 
Israel & 10 & New Zealand & 3 & Latvia & 1\\ 
Morocco & 10 & Mongolia & 3 & Singapore & 1\\ 
Lebanon & 10 & Ghana & 2 & Brunei Darussalam & 1\\ 
\hline
\end{tabular}%
}
\caption{Countries of origin in the \textbf{Russian} subset of \benchmarkname{}.}
\label{tab:ru_full}
\end{table*}

\begin{table*}[htbp]
\centering
\resizebox{\textwidth}{!}{%
\footnotesize
\begin{tabular}{ll | ll | ll}
\hline
\textbf{Country Name} & \textbf{Count} & \textbf{Country Name} & \textbf{Count} & \textbf{Country Name} & \textbf{Count} \\
\hline
Ukraine & 166 & Austria & 8 & Bolivia & 2\\ 
Russian Federation & 157 & Switzerland & 8 & Peru & 2\\ 
Belarus & 100 & Kyrgyzstan & 8 & South Africa & 2\\ 
France & 59 & Sweden & 8 & Argentina & 2\\ 
Italy & 47 & Iran & 8 & Iceland & 2\\ 
Poland & 36 & Belgium & 7 & Iraq & 2\\ 
Türkiye & 34 & Egypt & 7 & Philippines & 2\\ 
Georgia & 33 & Czechia & 7 & Afghanistan & 2\\ 
United States & 30 & Croatia & 6 & Paraguay & 2\\ 
India & 29 & Latvia & 6 & Sri Lanka & 2\\ 
Germany & 27 & Palestine & 6 & Thailand & 2\\ 
Japan & 27 & Malaysia & 5 & Turkmenistan & 2\\ 
Indonesia & 24 & Bosnia and Herzegovina & 5 & Singapore & 2\\ 
Lithuania & 20 & Tunisia & 5 & Ecuador & 1\\ 
Uzbekistan & 19 & Denmark & 5 & Senegal & 1\\ 
Greece & 19 & Brazil & 5 & Niger & 1\\ 
Armenia & 18 & Nepal & 5 & Kenya & 1\\ 
Spain & 18 & Jordan & 5 & Bahrain & 1\\ 
Azerbaijan & 17 & Slovakia & 4 & Myanmar & 1\\ 
United Kingdom & 16 & Pakistan & 4 & Benin & 1\\ 
Bulgaria & 16 & Israel & 4 & Chile & 1\\ 
Hungary & 15 & Bhutan & 4 & Trinidad and Tobago & 1\\ 
China & 14 & Albania & 4 & Haiti & 1\\ 
Mexico & 14 & Libya & 4 & Monaco & 1\\ 
Romania & 14 & Norway & 4 & Sudan & 1\\ 
Syria & 11 & Taiwan & 3 & Saudi Arabia & 1\\ 
Nigeria & 11 & Ghana & 3 & Yemen & 1\\ 
Portugal & 11 & Viet Nam & 3 & Congo, The Democratic Republic of the & 1\\ 
Tajikistan & 10 & Finland & 3 & Angola & 1\\ 
Serbia & 10 & Morocco & 3 & Venezuela & 1\\ 
Moldova, Republic of & 10 & Uruguay & 3 & Canada & 1\\ 
Korea, Republic of & 10 & Slovenia & 3 & Cyprus & 1\\ 
Netherlands & 9 & Algeria & 3 & Mauritania & 1\\ 
Kazakhstan & 9 & Estonia & 3 & Unknown & 1\\ 
Lebanon & 9 & Mongolia & 3 & Australia & 1\\ 
Austria & 8 & North Macedonia & 3 & Montenegro & 1\\ 
\hline
\end{tabular}%
}
\caption{Countries of origin in the \textbf{Ukrainian} subset of \benchmarkname{}.}
\label{tab:uk_full}
\end{table*}

\end{document}